\documentclass[letterpaper]{article} 
\usepackage[preprint]{aaai2027}  
\usepackage[hyphens]{url}  
\usepackage{graphicx} 
\usepackage{natbib}  
\usepackage{caption} 
\usepackage{amsmath}
\usepackage{algorithm}
 \usepackage[table]{xcolor}
\usepackage{algorithmic}
\usepackage{newfloat}
\usepackage{multirow}
\usepackage{listings}
\usepackage{comment} 
\DeclareCaptionStyle{ruled}{labelfont=normalfont,labelsep=colon,strut=off}

\floatstyle{ruled}
\newfloat{listing}{tb}{lst}{}
\floatname{listing}{Listing}

\usepackage{booktabs}
\title{Question-Specific Knowledge Graphs for Efficient Visual Reasoning}

\author {
    Ting-Chih Chen\textsuperscript{\rm 1},
    Emile van Krieken,
    Shujian Yu,
    Filip Ilievski
}
\affiliations {
    Vrije Universiteit Amsterdam\\
    \textsuperscript{\rm 1}t.c.chen@vu.nl
}

\begin{document}

\maketitle

\begin{abstract}
Recent work in visual question answering has shown that vision-language models can exhibit strong reasoning capabilities by translating visual inputs into textual representations. The effectiveness of this translation depends on how well visual details are retained; models need to surface and align both explicit and implicit knowledge sufficient to support reasoning, without introducing spurious assumptions. Existing methods that leverage detailed image captions introduce visual details unrelated to the reasoning task, inflating input token counts and increasing computational cost. To address these challenges, we propose~\textbf{VisKG}, a reinforcement learning (RL) framework in which models learn to translate visual content into question-specific knowledge graph (KG) representations. This process filters out perceptual noise while preserving the entity-relation structure needed for chain-of-thought reasoning, following the principle of minimum sufficient information. To ensure stable RL post-training, VisKG adopts Group reward-Decoupled Normalization Policy Optimization (GDPO). In addition, we strengthen the supervision stage with negative rationale samples, exposing the model to incorrect reasoning paths before RL post-training. Experimental results across science, mathematics, and general visual understanding benchmarks show that VisKG achieves performance comparable to or better than baselines, while requiring fewer tokens than caption-based representations. Moreover, training VisKG with GDPO improves accuracy by 2\% over its GRPO-trained counterpart on average. These results suggest that KG representations are a promising approach for supporting multi-step reasoning and open up future work on adaptively selecting the most suitable representation for a given task.
\end{abstract}


\section{Introduction}
\begin{figure}[t]
    \centering
    \includegraphics[width=0.9\columnwidth]{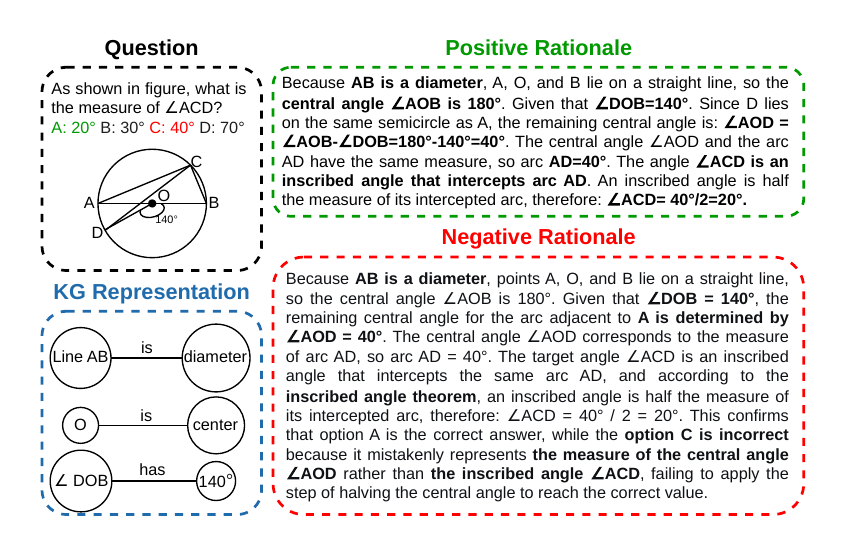}
    \caption{For a question-image VQA pair, VisKG generates a KG representation of the relevant visual content. Then, it produces KG-grounded CoT reasoning, illustrated with both a positive and a negative rationale.}
    \label{fig:1}
\end{figure}

Visual question answering (VQA) tasks require joint understanding of visual content and language to produce accurate answers. As questions become more complex, requiring multi-step inference, relational reasoning, or world knowledge~\cite{tran2025reasonvqa}, the challenge shifts from what a model knows to how effectively it integrates visual perception, linguistic understanding, and commonsense knowledge into coherent reasoning. For example, to answer the question in Figure~\ref{fig:1}, one needs to combine information from the question itself, visual information about the spatial configuration of the angles, and relevant mathematical background knowledge. Vision-language models (VLMs) have emerged as a paradigm for VQA tasks, leveraging world knowledge acquired through large-scale text corpora~\cite{zhang2024vision} to reason over questions that demand rich contextual understanding~\cite{chen-etal-2023-pre-trained}. However, visual content, linguistic input, and background knowledge are not expressed in a shared representation, making it difficult for VLMs to align the three into coherent reasoning~\cite{deng2026comprehensive}.

A natural approach to bridge vision and language is to convert visual inputs into textual representations, enabling VLMs to reason over linguistic inputs~\cite{wu-etal-2019-generating, 10677933}. A state-of-the-art (SOTA) method is Vision-SR1~\cite{li2026visionsr}, which aims to generate the image caption sufficient for answering the question without referring back to the image. To achieve this without external visual supervision, Vision-SR1 uses a two-rollout setting, decomposing reasoning into a visual and a language stage and applying a multi-reward RL objective that jointly optimizes the caption and the subsequent answer. However, captions are free-form text: they have no explicit mechanism for isolating which entities and relations are relevant to a given question, so irrelevant content cannot be selectively discarded. A knowledge graph (KG), by contrast, provides a structure in which visual entities and relations can be explicitly aligned with linguistic and background knowledge, while irrelevant nodes are pruned and the needed for reasoning is preserved~\cite{su-etal-2024-pipenet}.

We argue that effective RL post-training for VQA hinges on three interlocking design choices: the~\textit{representation} given to the model, the~\textit{algorithm} used to optimize it, and the~\textit{supervision signal} used to shape it. Existing RL-based VQA methods, including Vision-SR1, fall short on all three. First, on representation,~\textit{captions lack question-guided information selection}. Even when captions are optimized to align with the question, achieving this in free-form text is difficult in practice, often resulting in long captions that favor completeness over minimality. An ideal representation retains only what is necessary to answer the question, a principle known as minimum sufficient information~\cite{tishby2000information}. Yet, neither the caption format nor the prior training objectives enforce minimality, so captions tend toward completeness rather than question-specific relevance. Second, Group Relative Policy Optimization (GRPO)~\cite{deepseek-math} has become the de facto optimization algorithm for RL post-training, underlying many SOTA methods~\cite{shen2025vlm, zhu2026wikiseeker}. However,~\textit{GRPO is susceptible to reward collapse under heterogeneous advantage distributions}~\cite{gdpo}, undermining the stable convergence these methods depend on for reliable reasoning. Third, on the supervision signal, Supervised Fine-Tuning (SFT) provides only positive supervision,~\textit{exposing models to correct reasoning traces but not common reasoning errors}. Thus, models learn to reproduce correct patterns without understanding why plausible alternatives are wrong, leaving them prone to confident but flawed reasoning under distribution shift~\cite{tian2026learning}. These limitations compound one another: an uncontrolled representation exposes VLMs to task-irrelevant signals, an unstable algorithm prevents the models from learning reliably even from a good representation, and a positive supervision signal alone leaves the model without a signal for what to avoid, limiting its ability to generalize.

To address these limitations, we propose VisKG, an RL-based training framework for VQA that replaces image inputs with KG representations, as shown in Figure~\ref{fig:1}. Our contributions target each of the three design choices above: (1) on~\textit{representation}, we propose VisKG, a novel method to represent visual content as~\textit{question-specific KG} for reasoning; (2) on~\textit{algorithm}, we adopt a~\textit{GDPO-based}~\cite{gdpo} training formulation, deriving for the first time an optimization scheme that yields more stable RL optimization across diverse VQA tasks; and (3) on~\textit{supervision}, we enrich SFT training with negative rationales that expose VLMs to common reasoning failures before RL post-training, improving robustness across VQA benchmarks.

\section{Related Work}
\paragraph{Efficient Visual Representations for VQA.} Efficient visual representations aim to reduce redundancy by replacing raw pixels with language-aligned intermediates~\cite{teney2017graph, ziaeefard2020towards}: image/dense captioning generates textual scene descriptions~\cite{SALABERRIA2023118669, ge2024visual}, while region-level features and object lists extract salient semantic units~\cite{kim2019improving, hudson2019learning, khan2022expressive}, often further compressed via token-compression and visual-summarization techniques~\cite{chen2024efficient, lyu2025efficient, fang2026prune}. Yet, such task-agnostic summarization exposes models to redundant details, reinforcing shortcuts rather than grounding answers in question-relevant evidence~\cite{kanani2020improving, liu2025task}. VisKG instead uses KG representations optimized to capture minimal and sufficient reasoning signals, yielding a representation aligned with the evidence the question requires.

\paragraph{RL Post-Training in VLMs.} Recent work has adopted RL for post-training VLMs to align outputs with task-level rewards, reducing the reliance on expensive supervised signals and improving generalization to out-of-distribution tasks~\cite{tan2025interactive}. Vision-R1~\cite{xia2025visionary} addresses shortcut reliance by enforcing explicit image interpretation before reasoning via GRPO-based RL. VIDEOP2R~\cite{jiang2025videop2r} and Vision-SR1~\cite{li2026visionsr} factorize reasoning in the visual perception and language reasoning stages, using self-generated captions as a reward signal to reinforce both visual grounding and downstream reasoning. Within this caption-based RL paradigm, advantage estimation provides a relative quality signal for sampled outputs, letting the model distinguish higher- from lower-value trajectories and encouraging more deliberate, reward-aligned reasoning~\cite{park2023reinforcement}. Despite GRPO's promising results, its optimization dynamics can be sensitive to such variability. GDPO was recently introduced as a more robust alternative, improving stability through better variance control and baseline conditioning; however, it has only been explored in single-modality reasoning and remains unexamined for VQA. We adopt GDPO as VisKG's optimization framework to explore whether it enables stable and effective reasoning.

Before RL post-training, SFT remains mandatory to equip VLMs with the base capability needed for reward-driven optimization to be effective~\cite{yu2025reassessing}. However, existing SFT stage supervision focuses only on positive rationales, leaving models without explicit exposure to their own reasoning failures~\cite{ISLAM2022109083, nguyen2024diffusion, ding2026sherlock}. We complement positive with negative rationales in our SFT dataset for a stronger initialization before RL post-training.

\section{Methodology}
This section presents the VisKG framework for training VLMs: we describe the training procedure under a two-rollout setting (§\ref{VisKG_framework}), formalize the advantage score computation in GDPO (§\ref{pre}), introduce SFT preprocessing enriched with negative rationales (§\ref{pn}), and provide a theoretical interpretation grounded in the information bottleneck principle showing how the KG representations approximate minimally sufficient task information (§\ref{theory}).

\begin{figure*}[t]
    \centering
    \includegraphics[width=0.9\textwidth]{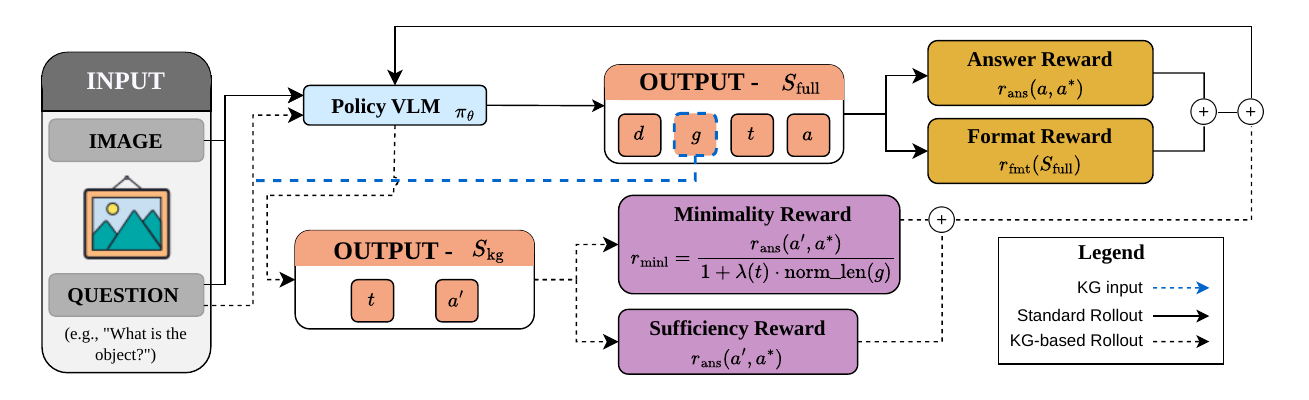}
    \caption{Overview of the VisKG training framework. The framework consists of two rollouts: standard rollout (black solid line) and KG-based rollout (black dashed line). In the standard rollout, the model generates information in a predefined format given an image and a question, and receives an answer reward and a format reward. In the KG-based rollout, the model generates a reasoning trace and final answer conditioned on the KG and the question, and receives a minimality reward and a sufficiency reward. The total reward is computed as the sum of all rewards from both rollouts and is used to update the VLM.}
    \label{fig:rl_rollout}
\end{figure*}

We consider a VLM $\pi_\theta$ that produces an answer $a$ to a question $q$ about an image $i$ by reasoning directly over the $i$. Rather than reasoning directly over $i$, our goal is to train $\pi_\theta$ to derive a question-specific KG $g$ that captures the information needed to answer $q$ without access to $i$, and to produce an answer $a'$ from $g$ instead. Concretely, $g$ should retain only the entities and relations relevant to $q$, discarding details that do not contribute to answering the $q$. This reflects the principle of minimum sufficient information: $g$ should be enough to omit task-irrelevant content, while still preserving the reasoning-critical structure required for $\pi_\theta$ to correctly predict $a'$, matching the ground-truth answer $a^*$.

\subsection{VisKG Training Framework}
\label{VisKG_framework}
We train the policy model $\pi_\theta$ to generate structured responses comprising four components: an image description $d$, a KG representation $g$, a CoT reasoning trace $t$, and an answer $a$. These components are serialized using markup tags to explicitly separate perceptual grounding, intermediate reasoning, and answer generation. For each training instance $\mathcal{Q}=\{i,q,a^*\}$, we perform two rollouts of the same $\pi_\theta$ within a single training step, as shown in Figure~\ref{fig:rl_rollout}.

\paragraph{Standard Rollout.}
The model receives the full multimodal input and generates a structured response 
$S_{\text{full}} \sim \pi_\theta(\cdot \mid i, q)$, where $S_{\text{full}} = (d,g,t,a)$. We evaluate both the correctness of the answers and the validity of the format. The~\textbf{answer reward} and~\textbf{format reward} are defined as:
\begin{equation}
\begin{aligned}
r_{\text{ans}}(a,a^*) &=
\begin{cases}
1, & a=a^*,\\
0, & \text{otherwise},
\end{cases}\\
r_{\text{fmt}}(S_\text{full}) &=
\begin{cases}
1, & \text{tags appear in order},\\
0, & \text{otherwise},
\end{cases}
\end{aligned}
\end{equation}
The answer reward follows the same design as Vision-SR1. However, the format reward differs: while Vision-SR1 only enforces a caption and reasoning structure, VisKG additionally requires a~\texttt{<KG>} tag between the description and reasoning stages, ensuring that the model explicitly grounds its reasoning in a structured KG representation. The standard rollout reward is then defined as: $r_{\text{full}} = r_{\text{ans}}(a,a^*) + r_{\text{fmt}}(S_\text{full}).$

\paragraph{KG-based Rollout.}
The KG $g$ is a subset of the set of all possible relational triples, $g\subseteq\{(e_i, r_k, e_j) \mid i, j \in \mathcal{I}_e,\ k \in \mathcal{I}_r\}$, where $\mathcal{I}_e$ is the index set over entities and $\mathcal{I}_r$ is the index set over relations. Here, $e_i, e_j$ denote visual or textual entities extracted from the image-question context, and $r_k$ represents a semantic relation between them. The model is then re-prompted with the generated KG $g$ (shown as the blue dashed block within $S_{\text{full}}$ in Figure~\ref{fig:rl_rollout}) together with the question $q$. While $g$ is a set of entity-relation triples, we serialize it as plain text rather than using a graph encoder. For example, a triple such as (angle A, has, angle B) is serialized as the plain-text string ``(angle A has angle B)'' before being tokenized, as illustrated in Figure~\ref{fig:1}. This produces a KG-based rollout $S_{kg} = (t, a') \sim \pi_\theta(\cdot \mid g, q)$, where $t$ is the CoT reasoning trace and $a'$ is the answer derived from $g$. If $a'$ is correct, the KG is considered sufficient for answering the question. We evaluate the KG along two dimensions:~\textbf{minimality}, which measures the compactness of $g$, and~\textbf{sufficiency}, which assesses whether $g$ preserves the information necessary for correct reasoning. Unlike Vision-SR1, which implicitly encourages sufficiency through self-reward but does not penalize verbosity, VisKG explicitly optimizes both dimensions as separate reward components. We operationalize these dimensions as reward components, as defined below.

\subparagraph{Minimality Reward.} KG complexity is normalized with respect to the mean KG length in the training corpus. Specifically, let $|g|$ denote the number of tokens in the serialized KG, i.e., when $g$ is linearized as a sequence of triples prior to being input to the model.
The normalized length is:
\begin{equation}
  \text{norm\_len}(g)
    = \frac{|g|}{L_{\text{ref}}},
  \qquad
  L_{\text{ref}}
    = \frac{1}{N}\sum_{j=1}^{N} |g_j|,
  \label{eq:norm_len}
\end{equation}
where $g_j$ is the serialized KG of the $j$-th sample in the SFT dataset (see §\ref{app:a1}), yielding $\text{norm\_len}(g)\!\approx\!1$ for average-length KGs. The minimality reward penalizes unnecessarily long KGs. With this penalty, the model can first learn to construct a valid KG before being pressured to compress it:
\begin{equation}
  r_{\text{minl}}
    = \frac{r_{\text{ans}}(a', a^*)}{1 + \lambda(t)\cdot\text{norm\_len}(g)},
\end{equation}
The constant 1 in the denominator ensures numerical stability when $\text{norm\_len}(g) = 0$; without it, an empty generated KG would cause the denominator to vanish, yielding an undefined reward signal. The penalty strength $\lambda(t)$ is annealed during training as:
\begin{equation}
  \label{eq:length_penalty}
  \lambda(t) = \lambda_{\max} \cdot \min\!\left(1,\,\frac{t}{T_{\text{warmup}}}\right),
\end{equation}
where $t$ is the current training step, $T_{\text{warmup}}$ is the number of warmup steps, and $\lambda_{\max}$ is the maximum value of the penalty coefficient $\lambda(t)$. $\lambda(t)$ increases linearly from $0$ to $\lambda_{\max}$ over the first $T_{\text{warmup}}$ steps so the model first learns to generate $g$ before the length penalty begins to shape its behavior.

\subparagraph{Sufficiency Reward.} The answer reward $r_{\text{ans}}$ acts as a sufficiency signal, directly assessing whether the generated KG preserves sufficient information to recover the answer $a^*$.

The KG-based reward is defined as $r_{\text{kg}} = r_{\text{minl}} + r_{\text{ans}}(a', a^*)$, where $a'$ denotes the answer predicted from the KG alone. The overall reward is $r_{\text{total}} = r_{\text{full}} + r_{\text{kg}}$.

\subsection{Advantage Computation in GDPO}
\label{pre}
Summing all reward components and normalizing the aggregated score, as in GRPO, is known to cause training instability in multi-reward settings: a single shared denominator lets the largest-scale reward dominate the pooled deviation, and negative covariance between components can suppress that same denominator toward zero, amplifying noise into disproportionately large advantages~\cite{gdpo}.
GDPO avoids both failure modes with~\emph{decoupled normalization}: each reward stream is normalized against its own statistics before being combined, then rescaled at the batch level. For a fixed prompt $i$ and reward $k$, the values across the $G$ rollouts in that group form a stream, normalized by
\begin{equation}
\begin{aligned}
A_k^{(i,j)} &= \frac{r_k^{(i,j)} - \mu(r_k^{(i,\cdot)})}{\sigma(r_k^{(i,\cdot)}) + \epsilon}, \quad
A_{\text{sum}}^{(i,j)} = \textstyle\sum_{k=1}^{n} A_k^{(i,j)}, \\
\hat{A}_{\text{sum}}^{(i,j)} &= \frac{A_{\text{sum}}^{(i,j)} - \mu_{\text{batch}}}{\sigma_{\text{batch}} + \epsilon},
\end{aligned}
\end{equation}
where $\mu(r_k^{(i,\cdot)})$ and $\sigma(r_k^{(i,\cdot)})$ are that stream's own mean and standard deviation, $\mu_{\text{batch}}$ and $\sigma_{\text{batch}}$ are computed over $A_{\text{sum}}^{(i,j)}$ across the mini-batch, and $\epsilon$ is a small constant guarding against near-zero variance in any single stream. Because each stream is normalized only by its own statistics, no component's scale can dominate another's, and cross-component cancellation can no longer suppress the shared denominator, resolving both failure modes above; a stream that is itself near-constant within a group (e.g., a format reward that is uniformly 1) is instead stabilized by the $\epsilon$ term. GDPO then rescales $A_{\text{sum}}^{(i,j)}$ at the batch level to keep the signal stable independent of $n$.
The resulting advantages are combined using a curriculum weight that gradually increases the influence of the KG-based reward as training progresses, reusing the same annealed penalty strength $\lambda(t)$ from Eq.~(\ref{eq:length_penalty}) so that the shift toward the KG-based advantage tracks the same warmup schedule that phases in the length penalty:
\begin{equation}
\label{cur}
\begin{aligned}
\tilde{A}_k &= w_{\text{full}}(t)\hat{A}^{\text{full}}_k
+ w_{\text{kg}}(t)\hat{A}^{\text{kg}}_k, \\
w_{\text{kg}}(t) &= \frac{\lambda(t)}{\lambda_{\max}}, \quad
w_{\text{full}}(t) = 1 - w_{\text{kg}}(t),
\end{aligned}
\end{equation}
where $\hat{A}^{\text{full}}_k$ and $\hat{A}^{\text{kg}}_k$ are the batch-normalized advantages, and $\tilde{A}_k$ is the advantage used in the policy update. $w_{\text{kg}}(t) = 0$ during the initial phase where the length penalty is inactive, and both mechanisms reach full strength together at $t = T_{\text{warmup}}$. Appendix~\ref{app:gdpo} shows the full procedure.

\subsection{SFT Preprocessing}
\label{pn}
Before RL training, we initialize $\pi_\theta$ with an SFT stage, following~\citet{chu2025sft}, who show that while RL generalizes better to out-of-distribution settings, SFT is still needed first to stabilize the model's output format, making it a necessary prerequisite for effective RL.

The SFT stage serves two objectives. The first is to instill compliance with VisKG's structured output format, ensuring well-formed responses before RL begins. The second, and a novel addition in our method, is to strengthen the model's reasoning prior by fine-tuning $\pi_\theta$ on both positive and negative rationale samples: positive rationales provide the correct reasoning trace leading to the right answer, while negative rationales identify why an incorrect answer fails and where the reasoning goes wrong. This supervision exposes the model to a wider range of reasoning trajectories than standard positive-only SFT regimes, which strengthens the reasoning prior for RL post-training. The SFT dataset construction are provided in §\ref{implement} and Appendix~\ref{app:a1}.

\subsection{Theoretical Interpretation}
\label{theory}
\begin{figure}[t]
    \centering
    \includegraphics[width=0.8\columnwidth]{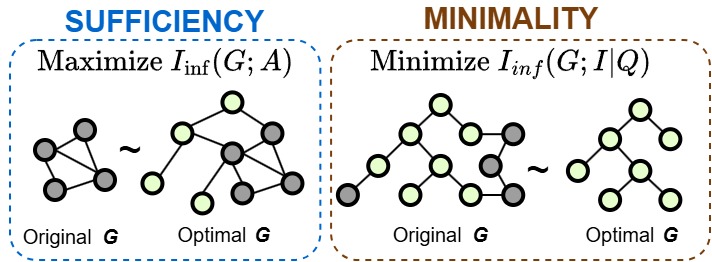}
    \caption{Illustration of the information bottleneck objective for KG generation. Sufficiency (left) maximizes $I_{\text{inf}}(G;A)$: the optimal $G$ retains nodes carrying mutual information with the answer (green), while gray nodes contribute little to prediction. Minimality (right) minimizes $I_{\text{inf}}(G;I\mid Q)$: $G$ discards gray nodes encoding question-irrelevant image information, keeping only the green nodes needed for reasoning.}
    \label{fig:ib}
\end{figure}

The objective used in our framework is to learn to generate a KG that is both~\textbf{sufficient} to answer the question and~\textbf{minimal} in the information it requires. Our solution can be interpreted from an information bottleneck (IB) perspective~\cite{tishby2000information}, which provides a principled penalty that jointly encourages sufficiency and minimality:
\begin{equation}
g^* \in \operatorname*{arg\,max}_{G} \; I_{\text{inf}}(G; A \mid Q) - \beta_{\text{IB}} \cdot 
I_{\text{inf}}(G; I \mid Q),
\end{equation}
where $\beta_{\text{IB}} > 0$ is the regularization coefficient, $g^*$ is the minimal sufficient KG representation that preserves the reasoning-critical content required for correct answer prediction, $I$ denotes the image, $Q$ the question, $G$ the KG, $A$ the answer, and $I_{\text{inf}}(\cdot\,;\cdot)$ denotes mutual information. The first term, $I_{\text{inf}}(G; A \mid Q)$, captures~\textbf{sufficiency}, requiring $G$ to retain the information necessary to predict the correct answer. In practice, this is approximated by the sufficiency reward from the KG-based rollout. The second term, $I_{\text{inf}}(G; I \mid Q)$, enforces~\textbf{minimality}, encouraging $G$ to discard visual information that is not required once the answer is determined, given the question. We approximate this term using the minimality reward from the KG-based rollout, which penalizes KG length as a proxy for unneeded information: a longer KG is more likely to encode visual content beyond what the question requires, as illustrated in Figure~\ref{fig:ib}.

\section{Experiments}
\subsection{Implementation Details}
\label{implement}
We construct a novel~\textit{PN-Rationales} dataset derived from a randomly selected 9K subset of the Vision-SR1-47K dataset~\cite{li2026visionsr}, covering three domains: general visual reasoning, mathematics, and scientific knowledge. For each sample, we prompt Qwen3-VL-32B-Instruct~\cite{qwen3technicalreport} with structured prompts to generate an image description, a question-specific KG, and a CoT rationale for each answer option, with InternVL3-78B~\cite{chen2024internvl} serving as a verifier to determine answer correctness and ensure data quality. For a sample with $k$ answer options, this yields a positive rationale, corresponding to the CoT for the correct option, and $k-1$ negative rationales, corresponding to the CoT for each incorrect option. 

To implement VisKG, we adopt Qwen2.5-VL-7B-Instruct~\cite{qwen2.5-VL} and Qwen3-VL-8B-Instruct~\cite{qwen3technicalreport} as base models. 
We first train the model via SFT on the~\textit{PN-Rationales} dataset. After SFT training, we optimize the model using GDPO for one epoch on the full Vision-SR1-47K dataset~\cite{li2026visionsr}, during which the policy model sequentially generates an image description, KG, CoT reasoning, and the final answer. Reusing Vision-SR1-47K enables direct comparison with Vision-SR1. Training hyperparameters are in Appendix~\ref{app:hyperparameters}.

\subsection{Experimental Setup}
We evaluate VisKG in two domains: general visual understanding and multimodal mathematical reasoning. For general visual understanding, we use MMMU-Pro~\cite{yue2024mmmu} and the cognition subset of MME~\cite{fu2025mme}, both of which stress cross-modal reasoning over raw perceptual recognition. For mathematical reasoning, we use MathVerse~\cite{zhang2024mathverse}, whose diagram-centric, multi-variant design isolates genuine visual grounding from language priors. For MMMU-Pro and MathVerse, we report accuracy. For MME, we report the cognition score, defined as the sum of ACC and ACC+: ACC measures question-level correctness, while ACC+ requires both questions paired with an image to be answered correctly, penalizing inconsistent reasoning. Further evaluation details are in Appendix~\ref{app:benchmark}.

We compare VisKG against four baselines spanning distinct visual grounding strategies.~\textbf{Vision-R1}~\cite{vision-r1} applies group-based RL with answer-correctness rewards and no explicit grounding mechanism, serving as a grounding-free reference.~\textbf{Perception-R1}~\cite{yu2026perceptionr} extends this with pre-extracted visual annotations, representing annotation-based grounding under privileged supervision.~\textbf{Visionary-R1}~\cite{xia2025visionary} is a caption-based RL method producing outputs in a caption--reason--answer format.~\textbf{Vision-SR1}~\cite{li2026visionsr} is a two-rollout GRPO framework trained on Vision-SR1-47K, where the second rollout conditions on a generated image description.

\subsection{Results and Discussion}

\begin{table}[t]
\centering
\scriptsize
\setlength{\tabcolsep}{3pt}
\resizebox{\linewidth}{!}{%
\begin{tabular}{l l c c}
\toprule
\multicolumn{4}{l}{\cellcolor{gray!30}\textit{Baselines}} \\
\midrule
\textbf{Method} & \textbf{Backbone} & \textbf{MMMU-Pro} & \textbf{MathVerse} \\
\midrule
Vision-R1     & Qwen2.5-VL-3B        & 34.9 & \textbf{57.3} \\
Visionary-R1  & Qwen2.5-VL-7B        & 27.4 & 45.0 \\
Perception-R1 & Qwen2.5-VL-3B-Inst.  & 36.8 & 52.1 \\
\cmidrule(lr){1-4}
\multirow{2}{*}{Vision-SR1}    & Qwen2.5-VL-3B        & \textbf{40.8} & 45.8 \\
              & Qwen2.5-VL-7B        & 40.7 & 54.5 \\
\midrule
\multicolumn{4}{l}{\cellcolor{gray!30}\textit{KG-based rollout inference}} \\
\midrule
\textbf{Backbone} & \textbf{Setting} & \textbf{MMMU-Pro} & \textbf{MathVerse} \\
\midrule
\multirow{4}{*}{Qwen2.5-VL-7B-Inst.}
  & ZS / ZS + GDPO          & 24.5 / 28.3            & 37.9 / 40.8 \\
  & SFT (n/p/b)             & 27.8 / 32.6 / 33.4     & 37.7 / 41.3 / 43.7 \\
  & SFT (n/p/b) + GRPO      & 33.8 / 37.1 / 38.6     & 43.9 / 47.8 / 54.1 \\
  & VisKG (SFT (n/p/b) + GDPO)      & 32.5 / 38.1 / \textbf{38.7} & 47.1 / 53.8 / \textbf{54.2} \\
\cmidrule(lr){1-4}
\multirow{4}{*}{Qwen3-VL-8B-Inst.}
  & ZS / ZS + GDPO          & 29.7 / 30.8            & 46.1 / 48.5 \\
  & SFT (n/p/b)             & 29.6 / 31.4 / 33.4     & 42.4 / 46.3 / 45.9 \\
  & SFT (n/p/b) + GRPO      & 35.2 / 36.1 / 37.4     & 46.0 / 51.2 / 52.1 \\
  & VisKG (SFT (n/p/b) + GDPO)      & 37.2 / \textbf{39.7} / 39.1 & 48.1 / \textbf{54.2} / 53.7 \\
\bottomrule
\end{tabular}%
}
\caption{Performance under different settings on MMMU-Pro and MathVerse. ZS denotes zero-shot inference. n, p, and b represent SFT using negative rationale samples, positive rationale samples, and both negative and positive rationale samples, respectively.}
\label{tab:MMMU-pro-mathverse-2-rollout}
\end{table}

\begin{table*}[!t]
\centering
\tiny
\setlength{\tabcolsep}{4pt}
\renewcommand{\arraystretch}{0.8}
\resizebox{\linewidth}{!}{
\begin{tabular}{l l c c c c c c c c c}
\toprule
\textbf{Backbone} & \textbf{Setting} 
& \multicolumn{2}{c}{\textbf{Commonsense Reasoning}}
& \multicolumn{2}{c}{\textbf{Numerical Calculation}}
& \multicolumn{2}{c}{\textbf{Text Translation}}
& \multicolumn{2}{c}{\textbf{Code Reasoning}}
& \textbf{Cog. Score} \\
\cmidrule(lr){3-4} \cmidrule(lr){5-6} \cmidrule(lr){7-8} \cmidrule(lr){9-10}
& & ACC & ACC+ & ACC & ACC+ & ACC & ACC+ & ACC & ACC+ & \\
\midrule
\multicolumn{11}{l}{\cellcolor{gray!30}\textit{KG-based rollout inference}} \\
\midrule
Qwen2.5-VL-7B-Inst. & ZS & 47.14 & 20 & 85 & 75 & 77.5 & 55 & 47.5 & 35 & 442.14 \\
& ZS + GDPO & 47.5 & 19 & 82.5 & 73 & 77.5 & 51 & 47.5 & 32 & 430 \\
& SFT (n) & 61.43 & 22.86 & 72.5 & 45 & \textbf{95} & 90 & 57.5 & 15 & 459.29 \\
& SFT (p) & 62.14 & 25.71 & 82.5 & 65 & 87.5 & 75 & 60 & 20 & 477.85 \\
& SFT (b) & 64.29 & 32.86 & 82.5 & 70 & \textbf{95} & 90 & 57.5 & 20 & 512.15 \\
& SFT (n) + GRPO & 62.37 & 43.77 & 87.5 & 75 & 87.3 & 74 & 63 & 35 & 527.94 \\
& SFT (p) + GRPO & 69.22 & 42 & 80 & 65 & 93.2 & \textbf{91} & 65.2 & 36 & 541.62 \\
& SFT (b) + GRPO & \textbf{70} & \textbf{49.12} & 90 & 85 & 88.2 & 83 & 71.23 & 50 & 586.55 \\
& VisKG (SFT (n) + GDPO) & 65.43 & 41.26 & 84 & 70 & 90 & 81 & 68 & 44 & 543.69 \\
& VisKG (SFT (p) + GDPO) & 63.19 & 34.71 & 92 & \textbf{88} & 92 & 90 & 67.3 & 39 & 566.2 \\
& VisKG (SFT (b) + GDPO) & 66.44 & 41.43 & \textbf{93} & 85 & 92.1 & 87 & \textbf{72.5} & \textbf{52} & \textbf{589.47} \\
\midrule
Qwen3-VL-8B-Inst. & ZS & 46.43 & 31.43 & 75 & 65 & 67.5 & 35 & 70 & 55 & 445.36 \\
& ZS + GDPO & 48 & 27 & 78.12 & 71 & 74.7 & 56 & 48.2 & 32 & 435.02 \\
& SFT (n) & 64.29 & 37.14 & 85 & 70 & 47.5 & 0 & 72.5 & 50 & 426.43 \\
& SFT (p) & 67.86 & 48.57 & 87.5 & 80 & 45 & 0 & 80 & 60 & 468.93 \\
& SFT (b) & 78.57 & 61.43 & \textbf{95} & \textbf{90} & 87.5 & 75 & 87.5 & 75 & 650 \\
& SFT (n) + GRPO & 63.55 & 36.71 & 80 & 60 & 42.5 & 20 & 62.8 & 38 & 403.56 \\
& SFT (p) + GRPO & 68.16 & 52.44 & 85 & 84 & 43 & 10 & 85 & 72 & 499.6 \\
& SFT (b) + GRPO & 74.98 & 65.1 & \textbf{95} & \textbf{90} & \textbf{92.5} & \textbf{85} & \textbf{92.5} & \textbf{85} & 680.08 \\
& VisKG (SFT (n) + GDPO) & 82.47 & 65.38 & 90 & 80 & 83.1 & 65 & 86.4 & 75 & 627.35 \\
& VisKG (SFT (p) + GDPO) & \textbf{83.1} & \textbf{72.4} & 88 & 85 & 82.8 & 67 & 91 & \textbf{85} & 654.3 \\
& VisKG (SFT (b) + GDPO) & 79.3 & 65.82 & \textbf{95} & \textbf{90} & 92.3 & \textbf{85} & 92.3 & \textbf{85} & \textbf{684.72} \\
\bottomrule
\end{tabular}}
\caption{Performance on the MME cognition subset under different training settings. ZS denotes zero-shot inference. n, p, and b denote SFT with negative rationales, positive rationales, and both negative and positive rationales, respectively.}
\label{tab:MME-2-rollout}
\end{table*}

\begin{figure*}[t]
    \centering
    \includegraphics[width=0.9\textwidth]{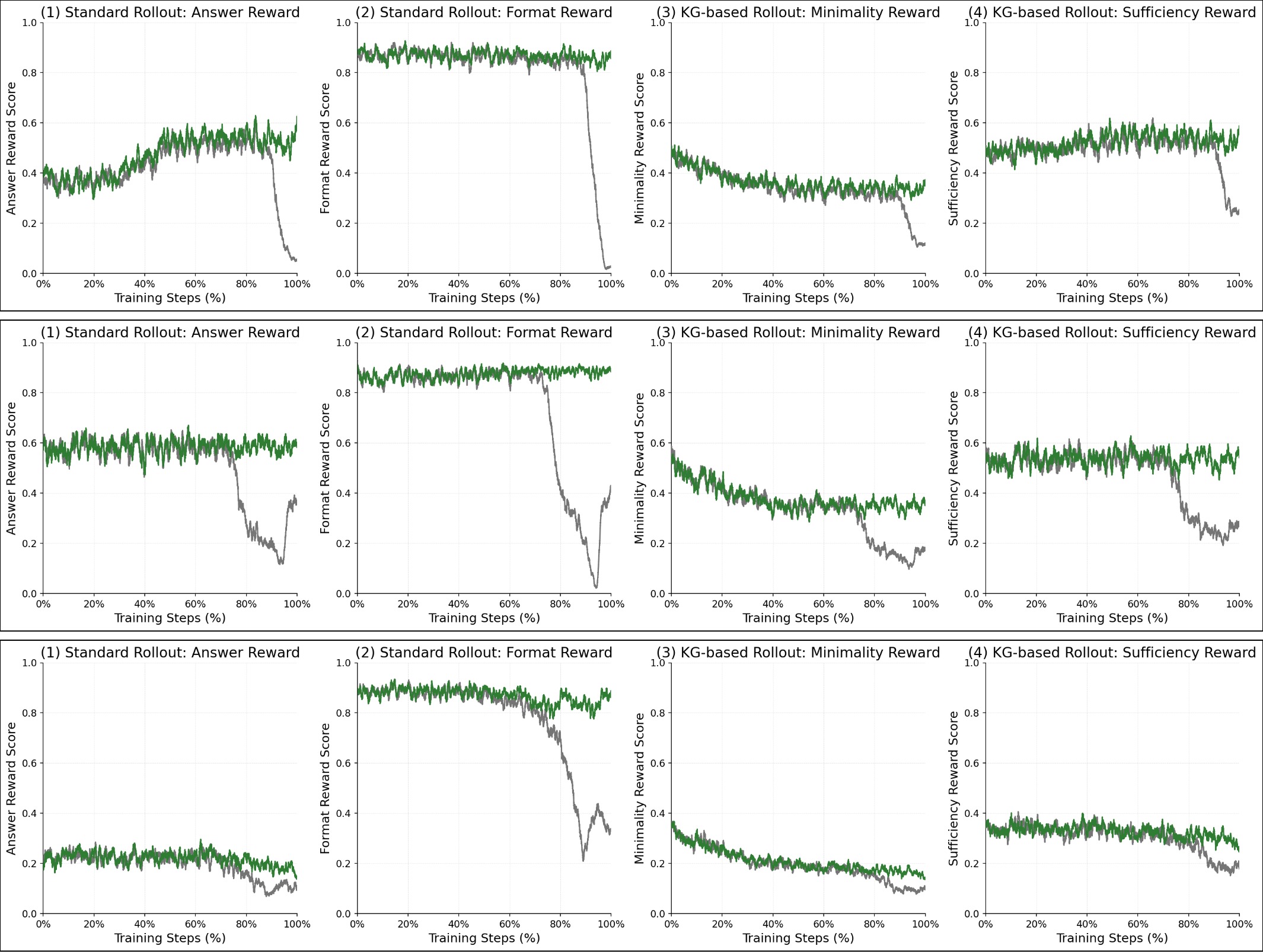}
    \caption{Training reward curves of Qwen2.5-VL-7B-Instruct under standard and KG-based rollouts. The green and gray lines correspond to GDPO and GRPO, respectively. Top, middle, and bottom panels show models initialized with SFT on both rationales, positive rationales only, and negative rationales only, respectively.}
    \label{fig:2.5_reward}
\end{figure*}

\subsubsection{How does the KG representation perform on VQA tasks compared to baselines?} As shown in Table~\ref{tab:MMMU-pro-mathverse-2-rollout}, VisKG achieves performance comparable to the baselines across both benchmarks. On MMMU-Pro, VisKG outperforms Vision-R1, Perception-R1, and Visionary-R1, and trails Vision-SR1 by only a small margin. A similar pattern holds on MathVerse, where VisKG remains competitive with Vision-R1 and Vision-SR1 despite using a more compact representation. Notably, VLMs can effectively answer VQA tasks using a variety of intermediate representations, though no single representation is universally optimal across domains; nevertheless, \textit{VisKG shows that a compact structure can match less compact alternatives while offering substantial efficiency gains.} We exclude MME from this comparison: the top-performing entries on the MME leaderboard are predominantly closed-source models with undisclosed architectures and training data, making any comparison against them uninformative rather than merely difficult. Restricting our evaluation to MMMU-Pro and MathVerse ensures that reported gains reflect genuine methodological differences rather than opaque, unreproducible advantages. Detailed results for the standard rollout and qualitative examples of VisKG inference are provided in Appendix~\ref{std_exp} and~\ref{VisKG_examples}, respectively.

\begin{table}[t]
\centering
\scriptsize
\setlength{\tabcolsep}{3pt}
\renewcommand{\arraystretch}{0.9}
\resizebox{\linewidth}{!}{%
\begin{tabular}{l l c c c c}
\toprule
\multicolumn{6}{l}{\cellcolor{gray!30}\textit{MMMU-Pro}} \\
\midrule
\textbf{Backbone} & \textbf{Setting} & \textbf{Acc. (Cap.)} & \textbf{Acc. (KG)} & \textbf{Avg. Cap. Len.} & \textbf{Avg. KG Len.} \\
\midrule
\multirow{3}{*}{Qwen2.5-VL-7B-Inst.}
  & SFT (n) + GDPO & 31.4 & \textbf{32.5} & 105 & 84  \\
  & SFT (p) + GDPO & 36.2 & \textbf{38.1} & 115 & 105 \\
  & SFT (b) + GDPO & 37.9 & \textbf{38.7} & 120 & 106 \\
\midrule
\multirow{3}{*}{Qwen3-VL-8B-Inst.}
  & SFT (n) + GDPO & 28.6 & \textbf{37.2} & 112 & 82  \\
  & SFT (p) + GDPO & \textbf{39.7} & \underline{39.7} & 114 & 99  \\
  & SFT (b) + GDPO & \textbf{39.4} & 39.1 & 116 & 102 \\
\midrule
\multicolumn{6}{l}{\cellcolor{gray!30}\textit{MathVerse}} \\
\midrule
\textbf{Backbone} & \textbf{Setting} & \textbf{Acc. (Cap.)} & \textbf{Acc. (KG)} & \textbf{Avg. Cap. Len.} & \textbf{Avg. KG Len.} \\
\midrule
\multirow{3}{*}{Qwen2.5-VL-7B-Inst.}
  & SFT (n) + GDPO & 45.3 & \textbf{47.1} & 127 & 92 \\
  & SFT (p) + GDPO & 47.5 & \textbf{53.8} & 120 & 112 \\
  & SFT (b) + GDPO & \textbf{54.8} & 54.2 & 140 & 106 \\
\midrule
\multirow{3}{*}{Qwen3-VL-8B-Inst.}
  & SFT (n) + GDPO & 46.7 & \textbf{48.1} & 108 & 96  \\
  & SFT (p) + GDPO & 52.4 & \textbf{54.2} & 107 & 105 \\
  & SFT (b) + GDPO & 53.1 & \textbf{53.7} & 123 & 100 \\
\bottomrule
\end{tabular}%
}
\caption{Inference cost evaluation on MMMU-Pro and MathVerse under the KG-based rollout setting. Avg. Cap. Len. and Avg. KG Len. denote the average length of the caption and KG intermediate representation, respectively.}
\label{tab:cost-2-rollout}
\end{table}

\subsubsection{What is the impact of SFT training on negative rationales on reasoning robustness?} We compare training settings and backbones. In Table~\ref{tab:MMMU-pro-mathverse-2-rollout}, Qwen2.5-VL-7B-Instruct consistently follows SFT(b) $>$ SFT(p) $>$ SFT(n) across SFT-only, GDPO, and GRPO. Qwen3-VL-8B-Instruct, however, breaks this ordering under GDPO on MMMU-Pro and under both GDPO and SFT-only on MathVerse. In Table~\ref{tab:MME-2-rollout}, both backbones consistently follow SFT(b) $>$ SFT(p) $>$ SFT(n) across all settings. These results suggest that~\textit{negative rationales improve reasoning robustness over positive-only supervision, though this effect weakens in larger, more reasoning-capable models.}

\subsubsection{To what extent does the IB-inspired reward formulation in GDPO improve reward stability during training relative to GRPO?} As shown in Table~\ref{tab:MMMU-pro-mathverse-2-rollout}, we compare VisKG's performance across different settings when trained with GDPO versus GRPO. GDPO-trained models outperform their GRPO-trained counterparts in nearly all settings, with only a single exception. Overall,~\textit{training VisKG with GDPO improves accuracy by 2 percentage points over its GRPO-trained counterpart on average.} A consistent pattern emerges in Table~\ref{tab:MME-2-rollout}, where every GDPO-trained setting outperforms its GRPO-trained counterpart. 

To further assess whether GDPO yields more stable RL training, we compare reward trajectories collected under the standard rollout and KG-based rollout for both GRPO and GDPO, as shown in Figure~\ref{fig:2.5_reward}. The results reveal a clear advantage for GDPO: models trained with GDPO update stably throughout the entire RL training process, whereas GRPO-trained rewards deteriorate sharply toward the end of training in most settings, with only a few configurations showing partial recovery. This comparison indicates:~\textit{GDPO produces more stable RL training than GRPO across both rollout settings.} We hypothesize that this stability gain stems from GDPO's decoupled normalization, which yields more distinct advantage scores during updates; we leave isolating this factor from GDPO's other design choices to future ablation studies. The training reward curves for Qwen3-VL-8B-Instruct are provided in Appendix~\ref{train_curve_3}.

\subsubsection{How does substituting KG representation with short image description in the SFT stage affect downstream task performance and input token efficiency?} To assess the necessity of the KG representation, we replace it with a short image caption, generated using the same method as the PN-Rationales dataset construction, during both the SFT and RL stages, keeping the rest of the VisKG framework unchanged. Table~\ref{tab:cost-2-rollout} reports results in the KG-based rollout setting. The accuracy remains comparable between the two representations, while KG is on average 16\% shorter in generated output length.~\textit{These findings suggest that the KG variant achieves performance comparable to a short caption while remaining a more efficient intermediate representation for reasoning.} Inference cost and KG question-specificity are analyzed in Appendix~\ref{cost} and~\ref{question-specific-KG}, respectively.

\section{Conclusion and Future Work}
We presented VisKG, an RL framework that trains VLMs to generate a question-specific KG representation as an intermediate reasoning structure for VQA. VisKG filters out irrelevant content and retains only the entity-relation structure necessary for reasoning. To improve reasoning robustness, we perform SFT on both positive and negative rationales. For RL post-training, VisKG adopts GDPO, marking its first application to multimodal VQA, which yields more stable training and a 2\% average accuracy gain over GRPO. VisKG achieves performance comparable to or better than baselines across science, mathematics, and general visual understanding benchmarks with fewer tokens than caption-based representations, establishing structured visual grounding as an efficient intermediate representation for VQA.

While VisKG demonstrates the effectiveness of KG-based representations for VQA, several limitations remain. First,~\textit{the downstream reasoning is not fully reliable}: VisKG focuses on generating a question-specific KG representation but does not explicitly govern how this representation is used during CoT reasoning, so answer quality still largely depends on the backbone VLM's own reasoning capability, and the same KG can yield very different performance across VLMs. Second,~\textit{VisKG is inherently biased toward entity-relation-centric questions}: since the KG mainly captures entities and their relationships, it offers limited benefit for VQA instances that do not follow this pattern. Future work will address both issues by explicitly coupling the KG with the reasoning process, e.g., via training objectives that condition CoT on specific KG components, and by exploring adaptive representation selection that chooses among KG, caption, or hybrid representations depending on the question type.

\section{Appendix}
\subsection{GDPO Optimization}
\label{app:gdpo}

For each input $Q = \{i, q\}$, where $i$ is the image and $q$ is the question, we sample $K$ candidate responses from the current policy:
\begin{equation}
  S_Q = \{s_1, \ldots, s_K\}, \quad s_k \sim \pi_\theta(\cdot \mid Q).
\end{equation}

Here, $s_k$ denotes the standard rollout response. For each $s_k$ we additionally generate a KG-conditioned response $s_{\text{kg},k} \sim \pi_\theta(\cdot \mid g_k, q)$, where $g_k$ is the KG produced during the standard rollout. 

\paragraph{Per-stream group-wise normalization.} Each standard rollout sample is evaluated under both reward streams, yielding $r_{\text{full},k}$ and $r_{\text{kg},k}$. We normalize each reward stream independently within the group of $K$ samples:
\begin{equation}
  \hat{A}^{\text{full}}_k
    = \frac{r_{\text{full},k} - \mu^{\text{full}}}{\sigma^{\text{full}}},
  \qquad
  \hat{A}^{\text{kg}}_k
    = \frac{r_{\text{kg},k} - \mu^{\text{kg}}}{\sigma^{\text{kg}}},
\end{equation}
where $\mu^{\cdot}$ and $\sigma^{\cdot}$ are the mean and standard deviation computed over the $K$ samples within the group. This decoupled normalization preserves fine-grained distinctions between reward combinations.

\paragraph{Curriculum-weighted combination.} The normalized advantages are combined using a curriculum weight that gradually increases the influence of the KG-based reward as training progresses. We reuse the same annealed penalty strength $\lambda(t)$ from Eq.~(\ref{eq:length_penalty}) as the driver of this curriculum, so that the shift toward the KG-based advantage tracks the same warmup schedule that phases in the length penalty:
\begin{equation}
\label{cur_app}
\begin{aligned}
\tilde{A}_k
&= w_{\text{full}}(t)\hat{A}^{\text{full}}_k
+ w_{\text{kg}}(t)\hat{A}^{\text{kg}}_k, \\
w_{\text{kg}}(t)
&= \frac{\lambda(t)}{\lambda_{\max}},
w_{\text{full}}(t) = 1 - w_{\text{kg}}(t),
\end{aligned}
\end{equation}
where the weights are set per training step according to
the $\lambda$-curriculum. The weights satisfy $w_{\text{full}}(t) + w_{\text{kg}}(t) = 1$
at every step.

\paragraph{Batch-wise advantage normalization.} Per-stream normalization ensures that the two reward signals are on a comparable scale before weighting; batch-wise normalization then stabilizes the magnitude of the final advantage regardless of how many reward streams are combined.
\begin{equation}
\begin{aligned}
A_k
&= \frac{\tilde{A}_k - \bar{\tilde{A}}}{\tilde{\sigma} + \varepsilon}, 
\bar{\tilde{A}}
= \frac{1}{K}\sum_{j=1}^{K}\tilde{A}_j, \\
\tilde{\sigma}
&= \sqrt{\frac{1}{K}\sum_{j=1}^{K}
\left(\tilde{A}_j - \bar{\tilde{A}}\right)^{2}},
\end{aligned}
\end{equation}
where $A_k$ is the final scalar advantage, $\tilde{A}_k$ is the curriculum-weighted advantage of the $k$-th, $K$ is the number of sampled responses per group, $\bar{\tilde{A}}$ is the group mean of $\tilde{A}_k$, $\tilde{\sigma}$ is the group standard deviation of $\tilde{A}_k$, and $\varepsilon$ is a small constant for numerical stability.

\paragraph{Policy objective.} We optimize the policy using the clipped surrogate objective from GDPO. Let $\pi_{\theta_{\text{old}}}$ denote the behavior policy used to sample $s_k$. The per-token probability ratio is:
\begin{equation}
  \rho_{k,\tau}(\theta)
    = \frac{\pi_\theta(s_{k,\tau} \mid Q,\, s_{k,<\tau})}
           {\pi_{\theta_{\text{old}}}(s_{k,\tau} \mid Q,\, s_{k,<\tau})},
\end{equation}
where $\tau$ indexes tokens in the response, $s_{k,<\tau}$ is the generated prefix up to step $\tau$, and $T_k = |s_k|$ is the response length. The per-sample clipped objective is:
\begin{equation}
\begin{aligned}
\mathcal{J}_k(\theta)
&= \frac{1}{T_k} \sum_{\tau=1}^{T_k}
\min\Bigl(
\rho_{k,\tau}(\theta)\,A_k,\;
\tilde{\rho}_{k,\tau}(\theta)\,A_k
\Bigr),
\\
\tilde{\rho}_{k,\tau}(\theta)
&=
\operatorname{clip}\bigl(\rho_{k,\tau}(\theta),
1-\varepsilon,\,1+\varepsilon\bigr),
\end{aligned}
\end{equation}
where $\varepsilon$ is the clipping threshold. A KL-divergence penalty is added to keep the updated policy close to a fixed reference policy $\pi_{\text{ref}}$, approximated as the difference in sequence log-probabilities:
\begin{equation}
  \widehat{\mathrm{KL}}_k
    = \log \pi_\theta(s_k \mid Q)
    - \log \pi_{\text{ref}}(s_k \mid Q),
\end{equation}
where the sequence log-probability is length-normalized:
\begin{equation}
  \log \pi_\theta(s_k \mid Q)
    = \frac{1}{T_k}
      \sum_{\tau=1}^{T_k}
      \log \pi_\theta(s_{k,\tau} \mid Q,\, s_{k,<\tau}).
\end{equation}

\paragraph{Final loss.} The per-sample loss combines the clipped surrogate objective with the
KL penalty:
\begin{equation}
  \mathcal{L}_k(\theta)
    = -\mathcal{J}_k(\theta)
    + \beta \cdot \widehat{\mathrm{KL}}_k,
\end{equation}
where $\beta$ controls the strength of KL regularization. The final objective averages over all $N = B \cdot K$ samples in the batch:
\begin{equation}
  \mathcal{L}_{\mathrm{GDPO}}(\theta)
    = \frac{1}{N}
      \sum_{b=1}^{B}\sum_{k=1}^{K}
      \Bigl[
        -\mathcal{J}_{b,k}(\theta)
        + \beta \cdot \widehat{\mathrm{KL}}_{b,k}
      \Bigr],
\end{equation}
where $B$ is the batch size and $K$ is the number of rollouts per input.

\paragraph{Parameter update.} Parameters are updated via gradient descent with gradient-norm clipping:
\begin{equation}
  \theta
    \leftarrow \theta
    - \eta_t \cdot
      \frac{\nabla_\theta \mathcal{L}_{\mathrm{GDPO}}(\theta)}
           {\max\!\left(1,\;
             \dfrac{\|\nabla_\theta \mathcal{L}_{\mathrm{GDPO}}(\theta)\|_2}
                   {g_{\max}}\right)},
\end{equation}
where $\eta_t$ is the learning rate at training step $t$ and $g_{\max}$ is the gradient-norm clipping threshold.

\begin{figure*}[t]
    \centering
    \includegraphics[width=0.9\textwidth]{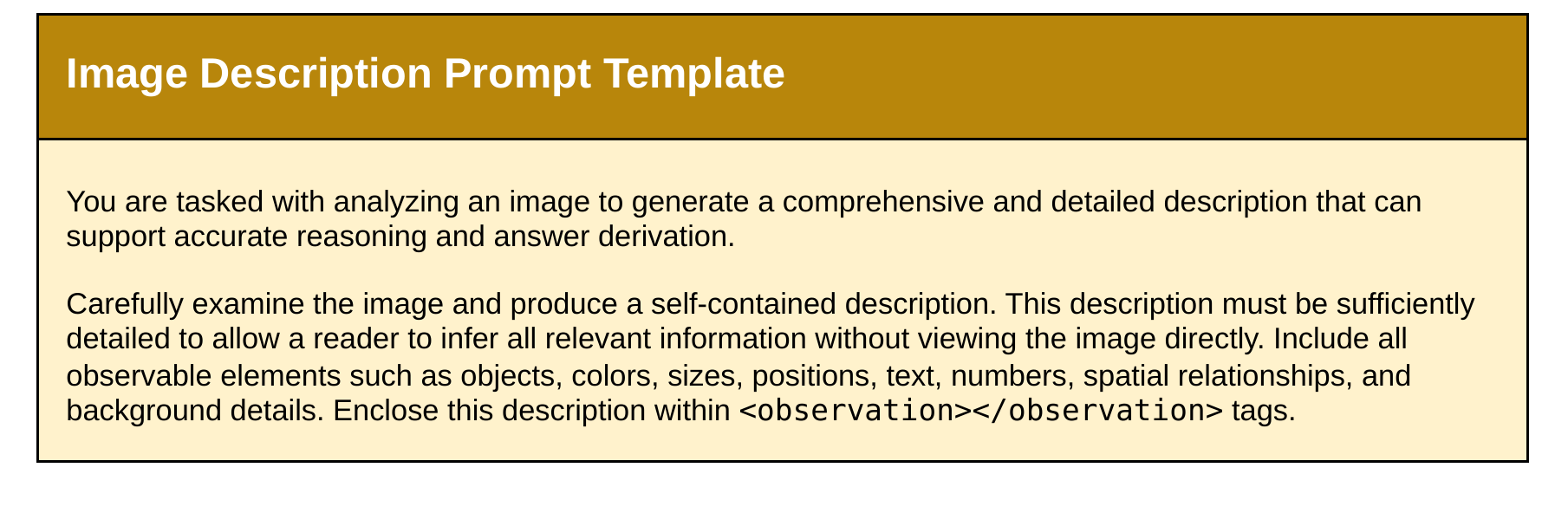}
    \includegraphics[width=0.9\textwidth]{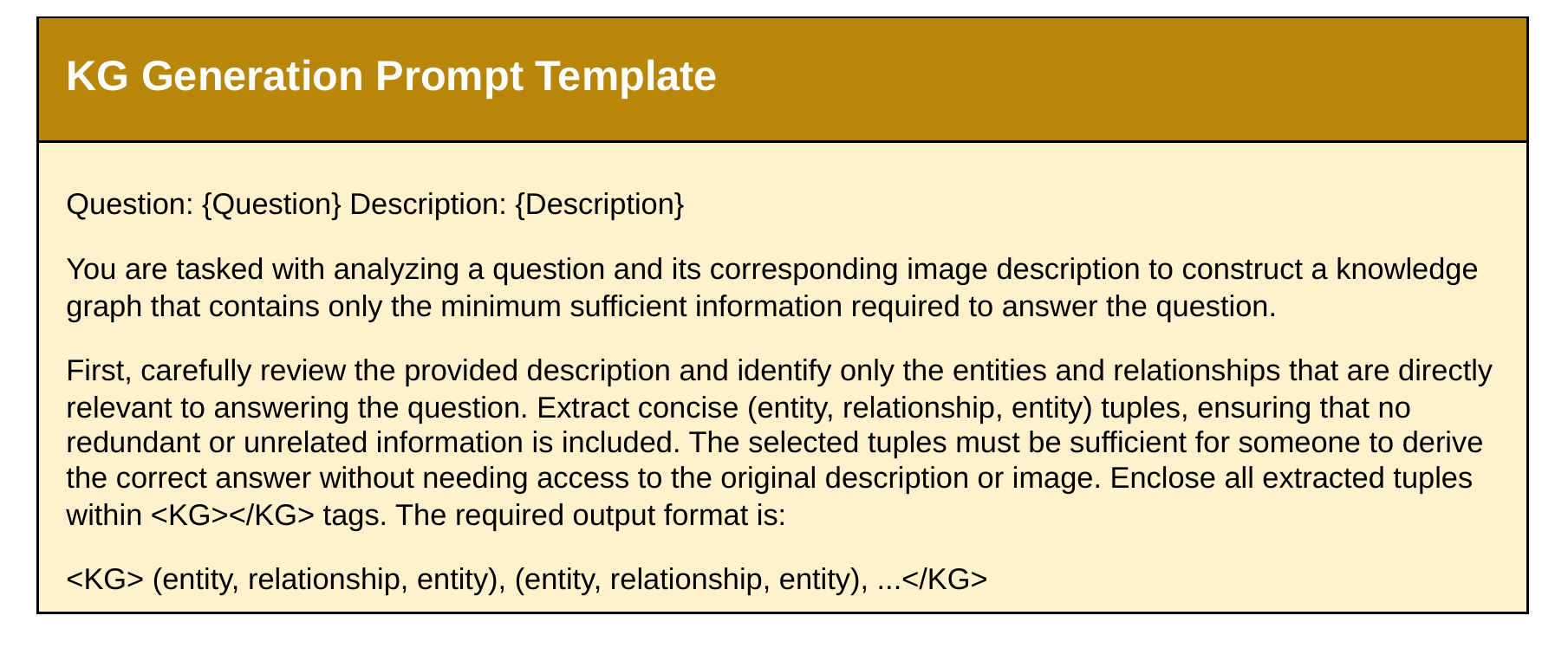}
    \caption{Prompts used to generate image description and KG.}
    \label{fig:prompts}
\end{figure*}

\begin{figure*}[t]
    \centering
    \includegraphics[width=0.9\textwidth]{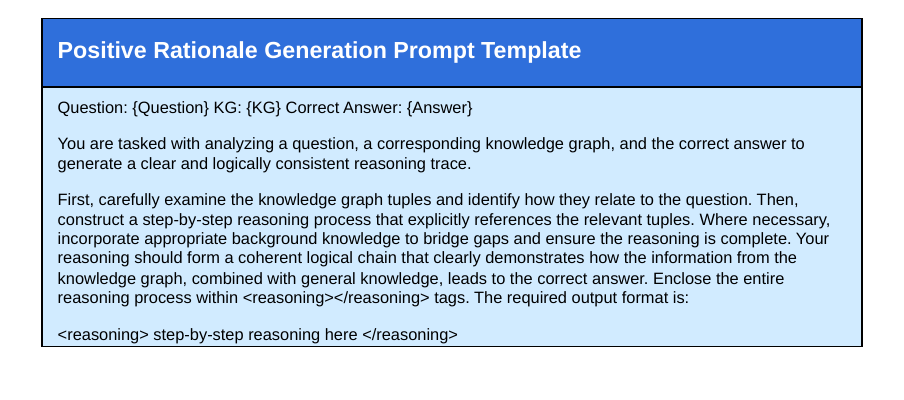}
    \includegraphics[width=0.9\textwidth]{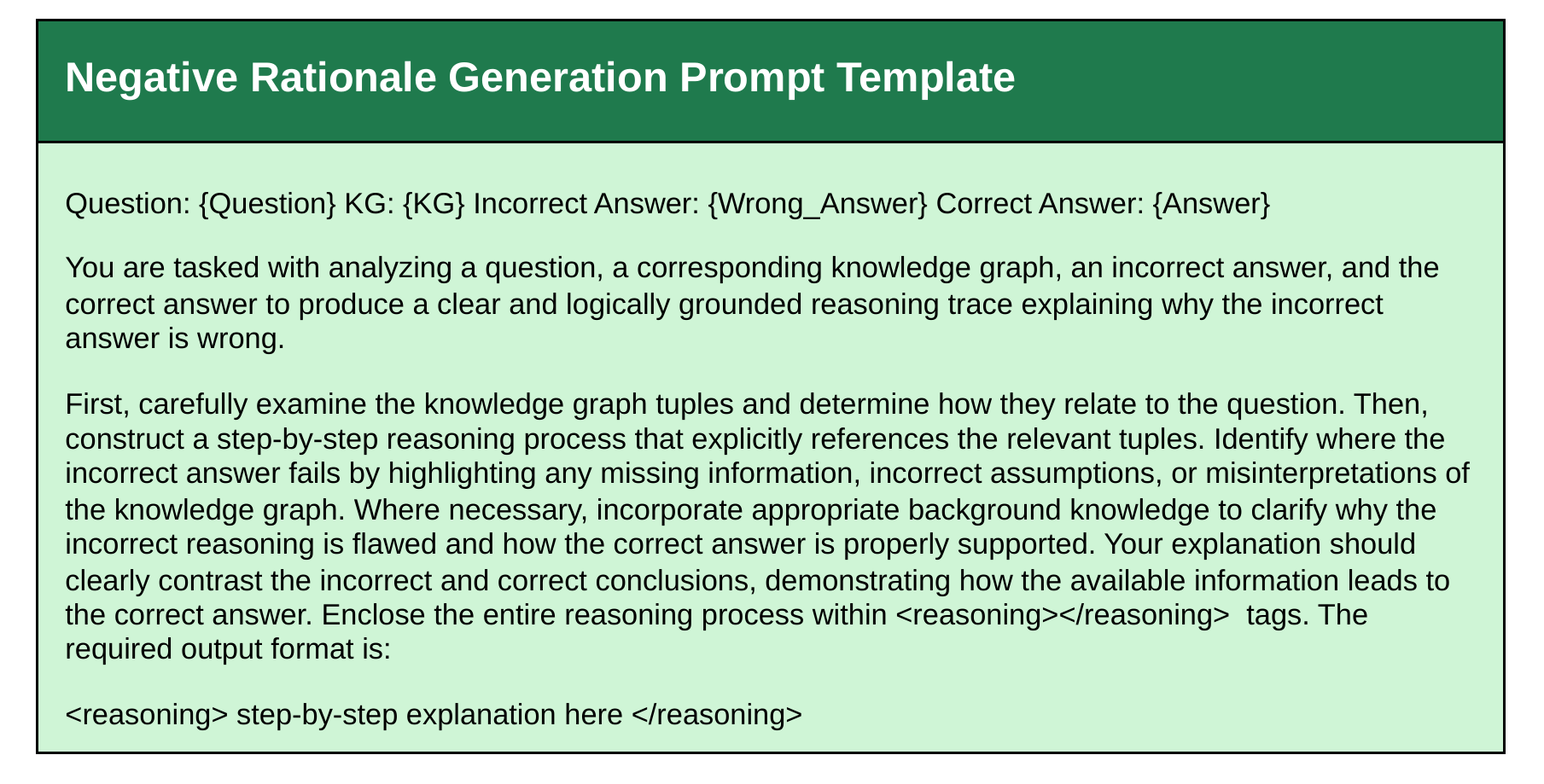}
    \caption{Prompts used to generate both positive and negative CoT rationales.}
    \label{fig:prompts_pos_neg}
\end{figure*}

\begin{figure*}[t]
    \centering
    \includegraphics[width=0.9\textwidth]{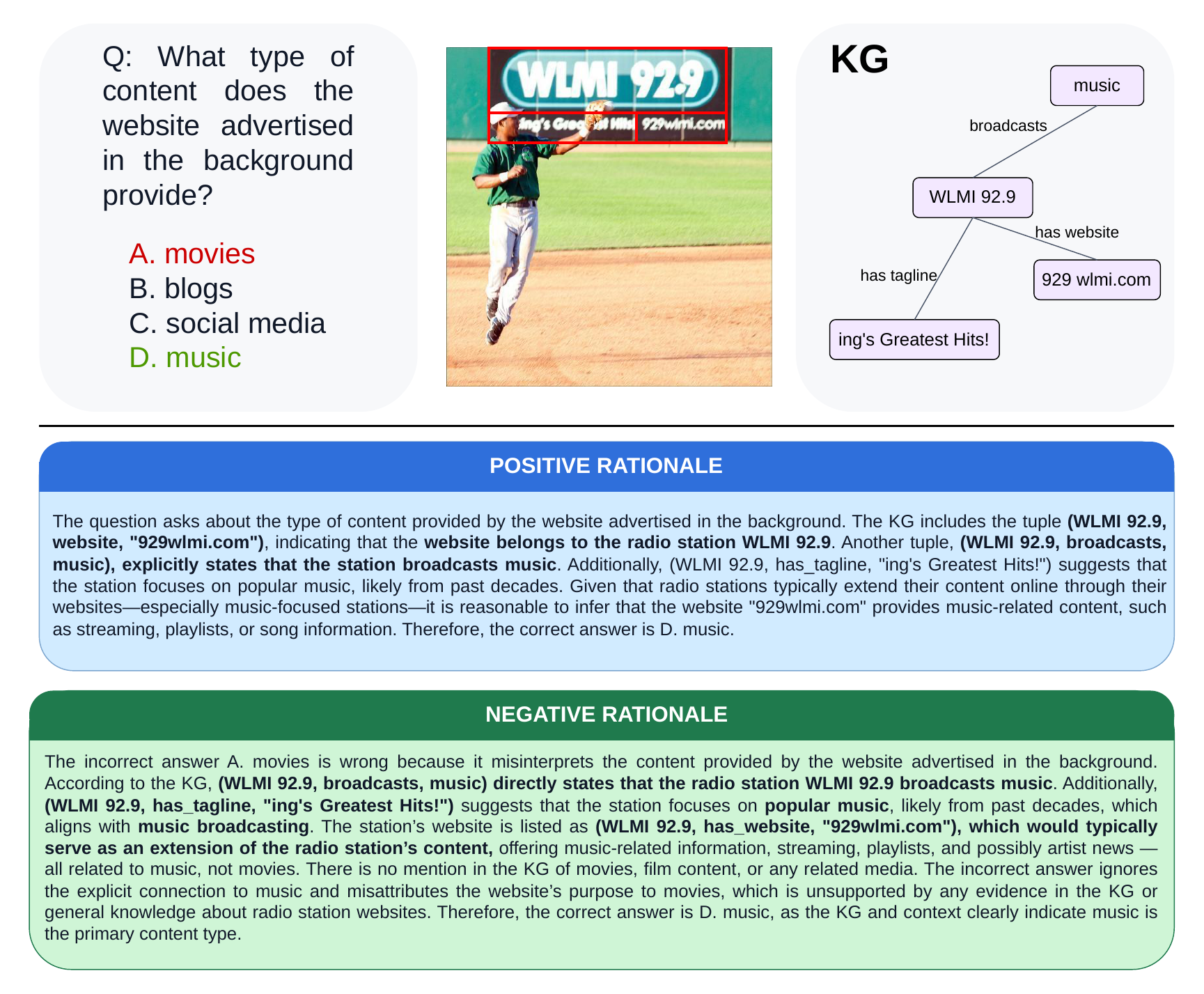}
    \caption{Positive and negative rationales derived from a question-specific KG. CoT reasonings are generated for both the correct answer and an incorrect answer.}
    \label{fig:sample}
\end{figure*}

\subsection{PN-Rationales Dataset}
\label{app:a1}

In this section, we present the prompts used to generate SFT training data for the PN-Rationales dataset and the detailed dataset statistics. All prompts are provided in Figure~\ref{fig:prompts} and~\ref{fig:prompts_pos_neg} for clarity and reproducibility. Figure~\ref{fig:sample} shows examples of positive and negative rationale samples. To construct high-quality supervision signals, we design structured prompts that guide the model to produce three types of outputs: (i) detailed image descriptions, (ii) question-specific KG, and (iii) both positive and negative CoT rationales. These prompts are carefully formulated to enforce consistency in format and to encourage the extraction of semantically meaningful and minimally sufficient information. The dataset statistics are presented in Table 4.

\begin{table}[t]
\label{tab:dataset_}
\centering
\small
\setlength{\tabcolsep}{4pt}
\renewcommand{\arraystretch}{1.1}
\begin{tabular}{l r r}
\toprule
\textbf{Source} & \textbf{Pos.} & \textbf{Neg.} \\
\midrule
\multicolumn{3}{l}{\textit{General Visual Reasoning}} \\
\midrule
TabMWP~\cite{lu2023dynamic} & 794 & 185 \\
OpenSpaces~\cite{chen2024spatialvlm} & 737 & 136 \\
Spacellava~\cite{foutter2025spacellavavisionlanguagemodeladapted} & 531 & 405 \\
ChartQA~\cite{masry2022chartqabenchmarkquestionanswering} & 263 & -- \\
A-OKVQA~\cite{a-okvqa} & 193 & 529 \\
PlotQA~\cite{Methani_2020_WACV} & 170 & -- \\
DVQA~\cite{kafle2018dvqaunderstandingdatavisualizations} & 159 & -- \\
FigureQA~\cite{kahou2018figureqaannotatedfiguredataset} & 99 & 101 \\
IconQA~\cite{lu2022iconqanewbenchmarkabstract} & 96 & 158 \\
MapQA~\cite{li2025mapqaopendomaingeospatialquestion} & 48 & 49 \\
\textbf{Subtotal} & \textbf{3,090} & \textbf{1,563} \\
\midrule
\multicolumn{3}{l}{\textit{Math}} \\
\midrule
Super-CLEVR~\cite{li2023super} & 384 & -- \\
UniGeo~\cite{chen-etal-2022-unigeo} & 366 & 1,002 \\
CLEVR-Math~\cite{arxiv.2208.05358} & 363 & -- \\
GeoQA+~\cite{cao-xiao-2022-augmented} & 225 & 610 \\
Geometry3K~\cite{lu2021inter} & 174 & 518 \\
GEOS~\cite{seo2015solving} & 14 & 42 \\
\textbf{Subtotal} & \textbf{1,526} & \textbf{2,172} \\
\midrule
\multicolumn{3}{l}{\textit{Science Knowledge}} \\
\midrule
PMC-VQA~\cite{zhang2023pmcvqa} & 257 & 752 \\
ScienceQA~\cite{lu2022learn} & 223 & 486 \\
AI2D~\cite{kembhavi2016diagram} & 187 & 601 \\
TQA~\cite{DBLP:conf/cvpr/KembhaviSSCFH17} & 171 & 501 \\
VQA-RAD~\cite{lau2018dataset} & 19 & 19 \\
\textbf{Subtotal} & \textbf{857} & \textbf{2,359} \\
\midrule
\textbf{Total} & \textbf{5,473} & \textbf{6,094} \\
\multicolumn{3}{r}{\textbf{Total samples: 11,567}} \\
\bottomrule
\end{tabular}
\caption{PN-Rationales dataset statistics. Pos.\ and Neg.\ denote the number of positive and negative rationale samples per source. ``--'' indicates the source does not appear in that split.}
\end{table}

\subsection{Training Hyperparameters}
\label{app:hyperparameters}

We train $\pi_\theta$ using LoRA with rank $r = 16$, scaling factor $\alpha_{\text{LoRA}} = 32$, and dropout $0.05$, applied on top of the SFT-initialized backbones. The RL optimizer uses a learning rate of $5 \times 10^{-7}$ with gradient-norm clipping at $g_{\max} = 1.0$ and mixed-precision training in \texttt{bf16}. For each training question, we sample $K = 4$ rollouts, with a maximum of $1{,}024$ new tokens per rollout for the standard rollout and KG-based rollout. The sampling is performed at temperature $0.7$. The batch size is $B = 20$ questions per step, and the training runs for $1$ epoch over the dataset. The KG length reference is set to $L_{\text{ref}} = 74.8$ tokens, pre-computed from the dataset. The length penalty is annealed via the $\lambda$-curriculum with $\lambda_{\max} = 1.0$ and a warmup ratio of $0.3$, i.e., $\lambda(t)$ reaches its maximum at $30\%$ of the total training steps. The KL regularization coefficient is $\beta = 0.1$, and the clipping threshold in the GDPO surrogate objective is $\varepsilon_{\text{clip}} = 0.2$. The reference policy $\pi_{\text{ref}}$ is the frozen SFT checkpoint, which is fixed throughout RL training. All experiments use~\texttt{flash\_attn} and gradient checkpointing, and are run on a single NVIDIA H100 (80GB) GPU with 1 node and 1 worker, with seed 42. Training one experimental setting takes approximately 4 days of wall-clock time.

\subsection{Benchmarks and Metrics Details}
\label{app:benchmark}

We evaluate model performance across two domains: (i) general visual understanding and (ii) multimodal mathematical reasoning.

\paragraph{General Visual Understanding.} We evaluate on~\textbf{MMMU-Pro}~\cite{yue2024mmmu} and~\textbf{MME (cognition)}~\cite{fu2025mme}. MMMU-Pro is a challenging benchmark that assesses cross-modal reasoning and subject-specific knowledge, presenting each question with ten candidate choices. In addition, it includes a vision-only setting in which all textual content is rendered directly within images, demanding robust visual comprehension in the absence of explicit linguistic cues. MME evaluates multimodal models across a broad spectrum of perception and cognition tasks, encompassing fine-grained visual understanding, optical character recognition, and commonsense reasoning, among others. We focus specifically on the \textit{cognition} subset, which targets higher-order capabilities such as logical inference and numerical reasoning, rather than the perception subset, which evaluates low-level visual recognition skills such as object existence and attribute detection. Since KG representations are designed to capture relational and semantic structure rather than raw perceptual details, the perception subset is less informative for evaluating the quality of KG-based reasoning. The cognition subset, by contrast, requires the model to integrate multiple visual cues simultaneously, providing a rigorous testbed for assessing whether the generated KG supplies structured evidence that is both sufficient and non-redundant.

\paragraph{Multimodal Mathematical Reasoning.} We evaluate on~\textbf{MathVerse}~\cite{zhang2024mathverse}, a benchmark comprising diagram-centric problems presented across six visual–textual variants. These variants are carefully designed to decouple reliance on visual inputs from language priors, thereby enabling a fine-grained assessment of whether a model's reasoning process is genuinely grounded in visual evidence rather than superficial textual patterns. This property makes MathVerse particularly well-suited to probing the practical utility of the KG generated by our method.

\paragraph{Evaluation Metrics.} For MMMU-Pro and MathVerse, we report accuracy as the primary evaluation metric. For MME, we report the cognition score, defined as the sum of ACC and ACC+. Each image in MME is paired with two binary (yes/no) questions. ACC measures performance at the question level, computing the proportion of individually correct responses. ACC+ is a stricter image-level metric, awarding credit only when~\textit{both} questions for a given image are answered correctly, thus penalizing inconsistent reasoning across question pairs. Together, ACC and ACC+ provide complementary views of model performance: ACC captures overall response correctness, while ACC+ assesses whether the model has genuinely understood the visual content rather than arriving at correct answers by chance.
\section{Ablation Experiments}

\subsection{Standard Rollout Experimental Results}
\label{std_exp}

We further conduct inference under the standard rollout setting. In Table~\ref{tab:MMMU-pro-mathverse-1-rollout}, standard rollout inference consistently outperforms KG-based rollout inference, and VisKG outperforms the baselines in several settings, most notably under SFT(b) + GDPO with Qwen3-VL-8B-Instruct. In Table~\ref{tab:MME-1-rollout}, the same pattern holds: standard rollout again outperforms KG-based rollout.

\subsection{VisKG Inference Examples}
\label{VisKG_examples}
\begin{figure*}[t]
    \centering
    \includegraphics[width=\textwidth]{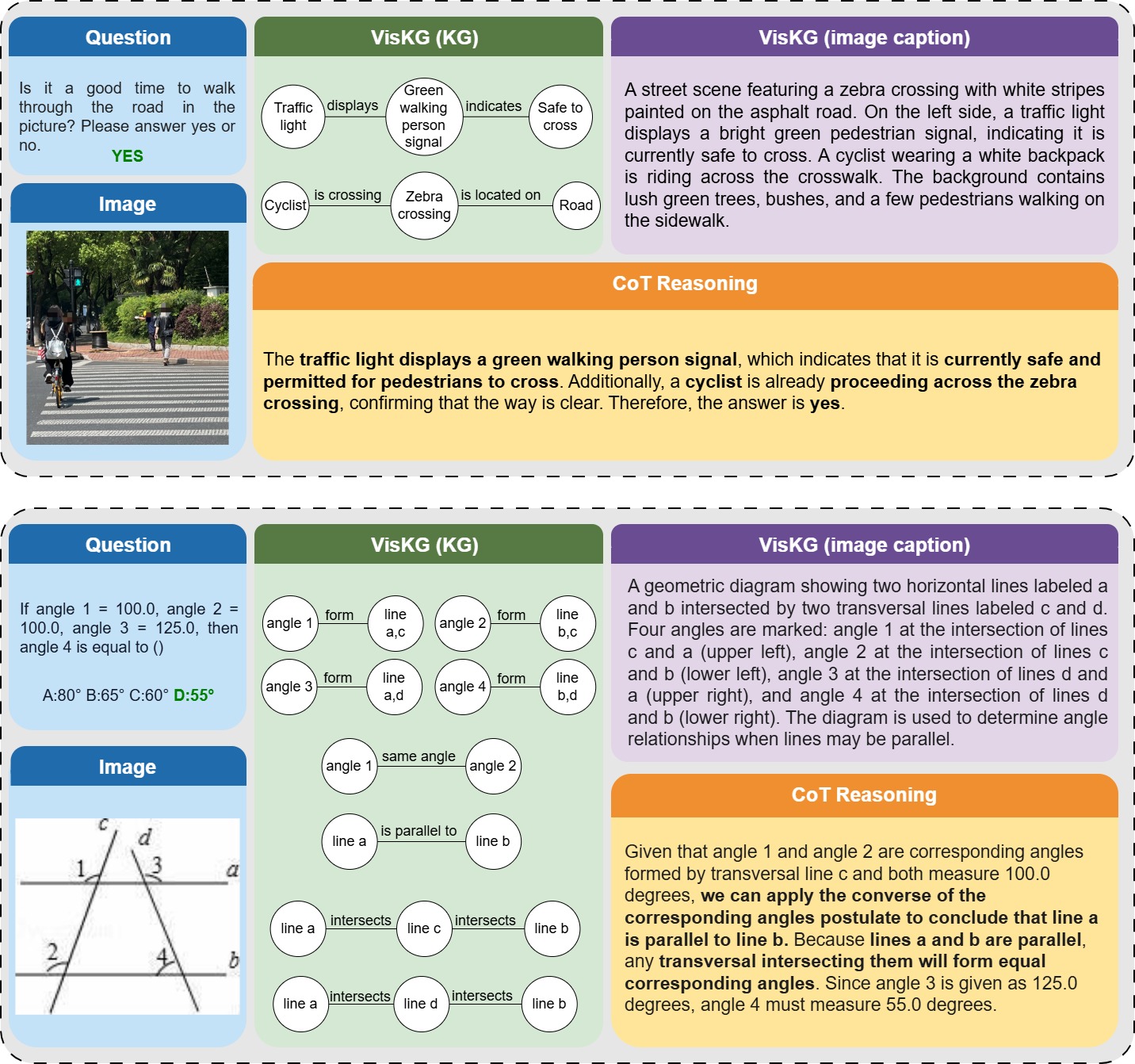}
    \caption{Qualitative examples of VisKG inference, comparing results obtained using the KG intermediate representation and the image caption. Each example shows the input image and question, the generated intermediate representation (KG and caption), and the resulting model answer.}
    \label{fig:inference_examples}
\end{figure*}

Figure~\ref{fig:inference_examples} shows qualitative examples of VisKG inference.

\begin{table}[ht]
\centering
\scriptsize
\setlength{\tabcolsep}{2.5pt}
\begin{tabular}{l l c c}
\toprule
\textbf{Method} & \textbf{Backbone} & \textbf{MMMU-Pro} & \textbf{MathVerse} \\
\midrule
\multicolumn{4}{l}{\cellcolor{gray!30}\textit{Baselines}} \\
\midrule
Vision-R1     & Qwen2.5-VL-3B        & 34.9 & \textbf{57.3} \\
Visionary-R1  & Qwen2.5-VL-7B        & 27.4 & 45.0 \\
Perception-R1 & Qwen2.5-VL-3B-Inst.  & 36.8 & 52.1 \\
\cmidrule(lr){1-4}
\multirow{2}{*}{Vision-SR1}    & Qwen2.5-VL-3B        & \textbf{40.8} & 45.8 \\
              & Qwen2.5-VL-7B        & 40.7 & 54.5 \\
\midrule
\multicolumn{4}{l}{\cellcolor{gray!30}\textit{Standard rollout inference}}\\
\midrule
Qwen2.5-VL-7B-Inst. & ZS / ZS + GDPO & 34.2 / 37.9 & 49.2 / 50.7 \\
& SFT (n/p/b) & 30.6 / 31.5 / 32.7 & 41.2 / 45.7 / 46.1\\
& SFT (n/p/b) + GRPO & 34.8 / 35.9 / 38.6 & 47.8 / 52.6 / \textbf{54.5} \\
& SFT (n/p/b) + GDPO & 35.9 / 39.7 / \textbf{40.1} & 48.7 / 54.2 / 54.3 \\
\cmidrule(lr){2-4}
Qwen3-VL-8B-Inst. & ZS / ZS + GDPO & 35.1 / 39.9 & 49.7 / 51.8 \\
& SFT (n/p/b) & 31.2 / 32.7 / 32.9 & 42.6 / 45.9 / 46.7 \\
& SFT (n/p/b) + GRPO & 33.8 / 36.9 / 38.6 & 48.9 / 49.2 / 54.3\\
& SFT (n/p/b) + GDPO & 35.1 / 39.7 / \textbf{41.7} & 48.7 / 54.1 / \textbf{54.8} \\
\toprule
\multicolumn{4}{l}{\cellcolor{gray!30}\textit{KG-based rollout inference}} \\
\midrule
\multirow{4}{*}{Qwen2.5-VL-7B-Inst.}
& ZS / ZS + GDPO & 24.5 / 28.3 & 37.9 / 40.8 \\
& SFT (n/p/b) & 27.8 / 32.6 / 33.4 & 37.7 / 41.3 / 43.7 \\
& SFT (n/p/b) + GRPO & 33.8 / 37.1 / 38.6 & 43.9 / 47.8 / 54.1 \\
& SFT (n/p/b) + GDPO & 32.5 / 38.1 / \textbf{38.7} & 47.1 / 53.8 / \textbf{54.2} \\
\cmidrule(lr){1-4}
\multirow{4}{*}{Qwen3-VL-8B-Inst.}
& ZS / ZS + GDPO & 29.7 / 30.8 & 46.1 / 48.5 \\
& SFT (n/p/b) & 29.6 / 31.4 / 33.4 & 42.4 / 46.3 / 45.9 \\
& SFT (n/p/b) + GRPO & 35.2 / 36.1 / 37.4 & 46.0 / 51.2 / 52.1 \\
& SFT (n/p/b) + GDPO & 37.2 / \textbf{39.7} / 39.1 & 48.1 / \textbf{54.2} / 53.7 \\
\bottomrule
\end{tabular}
\caption{Performance under different settings on MMMU-Pro and MathVerse. ZS denotes zero-shot inference. n, p, and b represent SFT using negative rationale samples, positive rationale samples, and both negative and positive rationale samples, respectively.}
\label{tab:MMMU-pro-mathverse-1-rollout}
\end{table}

\begin{table}[!htbp]
\centering
\scriptsize
\setlength{\tabcolsep}{3pt}
\renewcommand{\arraystretch}{0.9}
\resizebox{\linewidth}{!}{%
\begin{tabular}{l l c c c c}
\toprule
\textbf{Backbone} & \textbf{Method} &\textbf{Cog. Score (Cap.)} & \textbf{Cog. Score (KG)} & \textbf{Avg. Cap. Len.} & \textbf{Avg. KG Len.} \\
\midrule
\multirow{3}{*}{Qwen2.5-VL-7B-Inst.}
  & SFT (n) + GDPO & \textbf{544.78} & 543.69 & 70 & 52 \\
  & SFT (p) + GDPO & \textbf{568.92} & 566.2  & 56 & 57 \\
  & SFT (b) + GDPO & 585.28 & \textbf{589.47} & 89 & 62 \\
\midrule
\multirow{3}{*}{Qwen3-VL-8B-Inst.}
  & SFT (n) + GDPO & 621.94 & \textbf{627.35} & 67 & 50 \\
  & SFT (p) + GDPO & 649.48 & \textbf{654.3}  & 55 & 60 \\
  & SFT (b) + GDPO & 683.73 & \textbf{684.72} & 80 & 55 \\
\bottomrule
\end{tabular}%
}
\caption{Inference cost on the MME cognition subset under the KG-based rollout setting. Avg. Cap./KG Len.: average length of the generated caption/KG intermediate representation. Cog. Score (Cap./KG): cognition score obtained from the caption/KG representation.}
\label{tab:cost-MME-2-rollout}
\end{table}

\begin{table*}[!t]
\centering
\tiny
\setlength{\tabcolsep}{4pt}
\renewcommand{\arraystretch}{0.8}
\resizebox{\linewidth}{!}{
\begin{tabular}{l l c c c c c c c c c}
\toprule
\textbf{Backbone} & \textbf{Setting} 
& \multicolumn{2}{c}{\textbf{Commonsense Reasoning}}
& \multicolumn{2}{c}{\textbf{Numerical Calculation}}
& \multicolumn{2}{c}{\textbf{Text Translation}}
& \multicolumn{2}{c}{\textbf{Code Reasoning}}
& \textbf{Cog. Score} \\
\cmidrule(lr){3-4} \cmidrule(lr){5-6} \cmidrule(lr){7-8} \cmidrule(lr){9-10}
& & ACC & ACC+ & ACC & ACC+ & ACC & ACC+ & ACC & ACC+ & \\
\midrule
\multicolumn{11}{l}{\cellcolor{gray!30}\textit{Standard rollout inference}} \\
\midrule
Qwen2.5-VL-7B-Inst. & ZS & 80 & 60 & 77.5 & 55 & \textbf{95} & \textbf{90} & \textbf{87.5} & 75 & 620 \\
& ZS+GDPO & \textbf{80.71} & 64.29 & \textbf{95} & \textbf{90} & 75 & 55 & \textbf{87.5} & \textbf{80} & \textbf{627.5} \\
& SFT (n) & 78.54 & 57.2 & 73.2 & 47 & 92 & 84 & 87.1 & 71 & 590.04 \\
& SFT (p) & 79.29 & 58.57 & 75 & 50 & \textbf{95} & \textbf{90} & \textbf{87.5} & 75 & 610.36 \\
& SFT (b) & 80 & 60 & 75 & 50 & \textbf{95} & \textbf{90} & \textbf{87.5} & 75 & 612.5 \\
& SFT (n) + GRPO & 52.4 & 7.6 & 65 & 43 & 52.5 & 5 & 55 & 17 & 297.5 \\
& SFT (p) + GRPO & 75 & 60 & 60 & 27 & 62.5 & 29 & 52.5 & 20 & 386 \\
& SFT (b) + GRPO & 80.2 & \textbf{68.3} & 92.5 & 85 & 85 & 70 & 67.5 & 40 & 588.5 \\
& SFT (n) + GDPO & 80 & 60 & 77.5 & 54 & \textbf{95} & 87 & \textbf{87.5} & 71 & 612 \\
& SFT (p) + GDPO & 80.28 & 59.4 & 77 & 50 & \textbf{95} & 88 & 85 & 76 & 610.68 \\
& SFT (b) + GDPO & 80.24 & 61 & 77.5 & 52 & 91 & \textbf{90} & \textbf{87.5} & 75 & 614.24 \\
\midrule
Qwen3-VL-8B-Inst. & ZS & 82.14 & 67.14 & 72.5 & 60 & \textbf{87.5} & \textbf{75} & 95 & 90 & 629.28 \\
& ZS+GDPO & 85 & 72.5 & 87.5 & \textbf{75} & \textbf{87.5} & \textbf{75} & 95 & 90 & 667.5 \\
& SFT (n) & 81 & 70.43 & 82 & 70 & 81 & 68 & 91.2 & 82 & 625.63 \\
& SFT (p) & 85 & 72.86 & 85 & 70 & 85 & 70 & 92.5 & 85 & 645.36 \\
& SFT (b) & 85 & 72.86 & 87.5 & \textbf{75} & 85 & 70 & 95 & 90 & 660.36 \\
& SFT (n) + GRPO & 80.65 & 64.8 & 70 & 60 & 20 & 0 & 33 & 17 & 345.45 \\
& SFT (p) + GRPO & 74.92 & 63.9 & 55 & 23.5 & 85 & 65 & 62 & 43 & 472.32 \\
& SFT (b) + GRPO & \textbf{86.38} & \textbf{75.61} & \textbf{90} & 74 & 85 & 70 & \textbf{97.5} & \textbf{95} & 673.49 \\
& SFT (n) + GDPO & 84 & 72.8 & 83.18 & 70.4 & 82 & 69 & 91.37 & 82.5 & 635.25 \\
& SFT (p) + GDPO & 85 & 73.67 & 86.38 & 72 & 86.2 & 73 & 93.78 & 83 & 653.03 \\
& SFT (b) + GDPO & 85 & 73.19 & 88.2 & \textbf{75} & 87.3 & \textbf{75} & 95 & 90 & \textbf{668.69} \\
\midrule
\multicolumn{11}{l}{\cellcolor{gray!30}\textit{KG-based rollout inference}} \\
\midrule
Qwen2.5-VL-7B-Inst. & ZS & 47.14 & 20 & 85 & 75 & 77.5 & 55 & 47.5 & 35 & 442.14 \\
& ZS + GDPO & 47.5 & 19 & 82.5 & 73 & 77.5 & 51 & 47.5 & 32 & 430 \\
& SFT (n) & 61.43 & 22.86 & 72.5 & 45 & \textbf{95} & 90 & 57.5 & 15 & 459.29 \\
& SFT (p) & 62.14 & 25.71 & 82.5 & 65 & 87.5 & 75 & 60 & 20 & 477.85 \\
& SFT (b) & 64.29 & 32.86 & 82.5 & 70 & \textbf{95} & 90 & 57.5 & 20 & 512.15 \\
& SFT (n) + GRPO & 62.37 & 43.77 & 87.5 & 75 & 87.3 & 74 & 63 & 35 & 527.94 \\
& SFT (p) + GRPO & 69.22 & 42 & 80 & 65 & 93.2 & \textbf{91} & 65.2 & 36 & 541.62 \\
& SFT (b) + GRPO & \textbf{70} & \textbf{49.12} & 90 & 85 & 88.2 & 83 & 71.23 & 50 & 586.55 \\
& VisKG (SFT (n) + GDPO) & 65.43 & 41.26 & 84 & 70 & 90 & 81 & 68 & 44 & 543.69 \\
& VisKG (SFT (p) + GDPO) & 63.19 & 34.71 & 92 & \textbf{88} & 92 & 90 & 67.3 & 39 & 566.2 \\
& VisKG (SFT (b) + GDPO) & 66.44 & 41.43 & \textbf{93} & 85 & 92.1 & 87 & \textbf{72.5} & \textbf{52} & \textbf{589.47} \\
\midrule
Qwen3-VL-8B-Inst. & ZS & 46.43 & 31.43 & 75 & 65 & 67.5 & 35 & 70 & 55 & 445.36 \\
& ZS + GDPO & 48 & 27 & 78.12 & 71 & 74.7 & 56 & 48.2 & 32 & 435.02 \\
& SFT (n) & 64.29 & 37.14 & 85 & 70 & 47.5 & 0 & 72.5 & 50 & 426.43 \\
& SFT (p) & 67.86 & 48.57 & 87.5 & 80 & 45 & 0 & 80 & 60 & 468.93 \\
& SFT (b) & 78.57 & 61.43 & \textbf{95} & \textbf{90} & 87.5 & 75 & 87.5 & 75 & 650 \\
& SFT (n) + GRPO & 63.55 & 36.71 & 80 & 60 & 42.5 & 20 & 62.8 & 38 & 403.56 \\
& SFT (p) + GRPO & 68.16 & 52.44 & 85 & 84 & 43 & 10 & 85 & 72 & 499.6 \\
& SFT (b) + GRPO & 74.98 & 65.1 & \textbf{95} & \textbf{90} & \textbf{92.5} & \textbf{85} & \textbf{92.5} & \textbf{85} & 680.08 \\
& VisKG (SFT (n) + GDPO) & 82.47 & 65.38 & 90 & 80 & 83.1 & 65 & 86.4 & 75 & 627.35 \\
& VisKG (SFT (p) + GDPO) & \textbf{83.1} & \textbf{72.4} & 88 & 85 & 82.8 & 67 & 91 & \textbf{85} & 654.3 \\
& VisKG (SFT (b) + GDPO) & 79.3 & 65.82 & \textbf{95} & \textbf{90} & 92.3 & \textbf{85} & 92.3 & \textbf{85} & \textbf{684.72} \\
\bottomrule
\end{tabular}}
\caption{Performance on the MME cognition subset under different training settings. ZS denotes zero-shot inference. n, p, and b denote SFT with negative rationales, positive rationales, and both negative and positive rationales, respectively.}
\label{tab:MME-1-rollout}
\end{table*}

\subsection{Training reward curves of Qwen3-VL-8B-Instruct}
\label{train_curve_3}

\begin{figure*}[t]
    \centering
    \includegraphics[width=\textwidth]{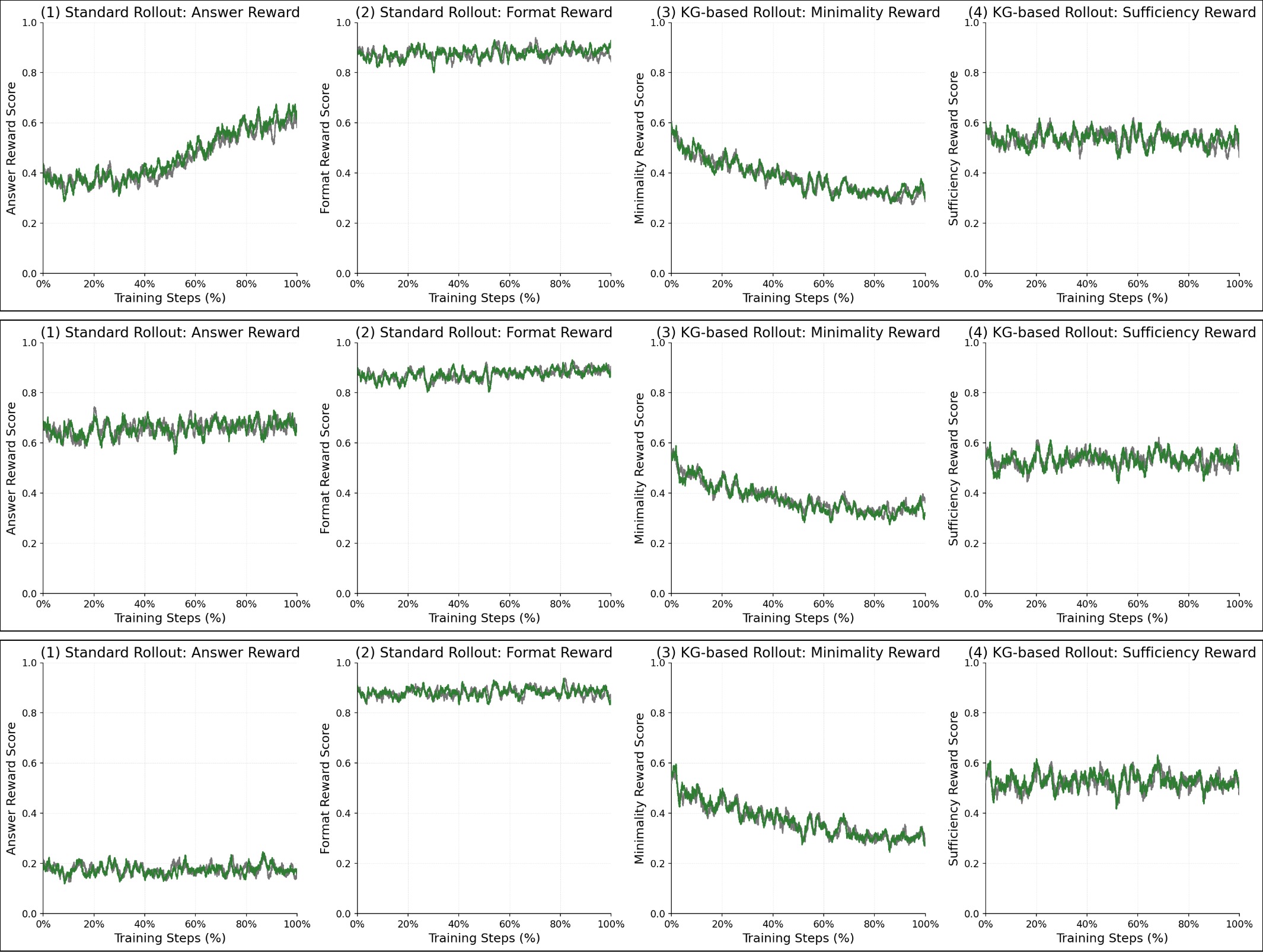}
    \caption{Training reward curves of Qwen3-VL-8B-Instruct under standard and KG-based rollouts. The green and gray lines correspond to GDPO and GRPO, respectively. Top, middle, and bottom panels show models initialized with SFT on both rationales, positive rationales only, and negative rationales only, respectively.}
    \label{fig:3_reward}
\end{figure*}

Reward trajectories collected under the standard rollout and KG-based rollout with GRPO and GDPO are further compared for Qwen3-VL-8B-Instruct, as shown in Figure~\ref{fig:3_reward}. This comparison reveals a different pattern from that observed in Figure~\ref{fig:2.5_reward}. For the Qwen2.5-VL-7B-Instruct backbone, GDPO exhibits more stable RL training than GRPO, since GRPO's rewards deteriorate toward the end of training. In contrast, the results in Figure~\ref{fig:3_reward} show that GRPO and GDPO achieve comparably stable RL training. We attribute this difference to the larger number of parameters in Qwen3-VL-8B-Instruct, which appears to mitigate the instability otherwise observed with GRPO. This observation is consistent with findings reported in~\cite{tan2025scaling}, suggesting that models with more parameters tend to exhibit more stable RL training. While GRPO's instability risk appears to diminish at larger parameter scales, GDPO maintains stable RL training across model scales, indicating that its stability does not depend on model size.

\begin{table}[ht]
\centering
\scriptsize
\setlength{\tabcolsep}{3pt}
\renewcommand{\arraystretch}{0.9}
\resizebox{\linewidth}{!}{%
\begin{tabular}{l l c c}
\toprule
\textbf{Backbone} & \textbf{Method} & \textbf{Triples Overlap Ratio (Avg.)} & \textbf{Cog. Score} \\
\midrule
\multirow{9}{*}{Qwen2.5-VL-7B-Inst.}
  & SFT (n)        & 29.68\% & 459.29 \\
  & SFT (n) + GRPO & 26.78\% & 527.94 \\
  & SFT (n) + GDPO & 23.19\% & \textbf{543.69} \\
  \cmidrule(lr){2-4}
  & SFT (p)        & 27.92\% & 477.85 \\
  & SFT (p) + GRPO & 18.40\% & 541.62 \\
  & SFT (p) + GDPO & 22.95\% & \textbf{566.2}  \\
  \cmidrule(lr){2-4}
  & SFT (b)        & 28.42\% & 512.15 \\
  & SFT (b) + GRPO & 22.81\% & 586.55 \\
  & SFT (b) + GDPO & 26.70\% & \textbf{589.47} \\
\midrule
\multirow{9}{*}{Qwen3-VL-8B-Inst.}
  & SFT (n)        & 27.41\% & 426.43 \\
  & SFT (n) + GRPO & 12.60\% & 403.56 \\
  & SFT (n) + GDPO & 18.39\% & \textbf{627.35} \\
  \cmidrule(lr){2-4}
  & SFT (p)        & 28.92\% & 468.93 \\
  & SFT (p) + GRPO & 26.44\% & 499.6  \\
  & SFT (p) + GDPO & 28.51\% & \textbf{654.3}  \\
  \cmidrule(lr){2-4}
  & SFT (b)        & 29.48\% & 650    \\
  & SFT (b) + GRPO & 29.10\% & 680.08 \\
  & SFT (b) + GDPO & 28.59\% & \textbf{684.72} \\
\bottomrule
\end{tabular}%
}
\caption{Triples overlap ratio and cognition score on the MME cognition subset, measuring the question-specificity of generated KG.}
\label{tab:mme-question-specific-kg}
\end{table}

\subsection{Cost Experiment}
\label{cost}
We further investigate the performance and cost of the VisKG training framework on the MME cognition subset, comparing inference using image captions versus KG representations, as shown in Table~\ref{tab:cost-MME-2-rollout}. The results show that cognition scores are similar between the two representations, with no clear trend favoring either intermediate representation. In terms of cost, however, the average caption length exceeds the average KG length by approximately 20\% in token count.

\subsection{Analysis of KG Question-Specificity}
\label{question-specific-KG}
Table~\ref{tab:mme-question-specific-kg} reports the triples overlap ratio and cognition score on the MME cognition subset, measuring how question-specific the generated KGs are. The overlap ratio is the proportion of shared triples between KGs generated from the same image under two different questions; a lower ratio indicates more question-specific generation. Across both backbones, SFT-only models consistently show higher overlap ratios and lower cognition scores than their GDPO- or GRPO-trained counterparts, indicating that SFT alone yields KGs with substantial redundant, question-agnostic content. RL training consistently lowers the overlap ratio while raising cognition score, showing that it pushes the model away from generic, overlapping KG content toward question-specific KGs that improve reasoning performance.

\bibliography{aaai2027}


\end{document}